\documentclass[pdflatex,sn-mathphys-num]{sn-jnl}%

\usepackage{algorithm}%
\usepackage{algorithmicx}%
\usepackage{algpseudocode}%
\usepackage{amsmath,amssymb,amsfonts}%
\usepackage{amsthm}%
\usepackage[title]{appendix}%
\usepackage{booktabs}%
\usepackage{graphicx}%
\usepackage{listings}%
\usepackage{makecell}%
\usepackage{manyfoot}%
\usepackage{mathrsfs}%
\usepackage{multirow}%
\usepackage{textcomp}%
\usepackage{xcolor}%
\usepackage{rotating}

\theoremstyle{thmstyleone}%

\theoremstyle{thmstyletwo}%

\theoremstyle{thmstylethree}%

\begin{document}

\title[Reducing Class Bias In Data-Balanced Datasets Through Hardness-Based Resampling]{Reducing Class Bias In Data-Balanced Datasets Through Hardness-Based Resampling}

\author[1]{Pawel Pukowski}\email{ppukowski1@sheffield.ac.uk}

\author*[2]{Venet Osmani}\email{v.osmani@qmul.ac.uk}

\affil[1]{Department of Computer Science, The University of Sheffield, United Kingdom}

\affil[2]{Digital Environment Research Institute, Queen Mary University of London, United Kingdom}

\abstract{Class-bias, that is class-wise performance disparities, is typically attributed to data imbalance and addressed through frequency-based resampling. However, we demonstrate that substantial bias persists even in perfectly balanced datasets, proving that class frequency alone cannot explain unequal model performance. We investigate these disparities through the lens of class-level learning difficulty and propose Hardness-Based Resampling (HBR), a strategy that leverages hardness estimates to guide data selection. To better capture these effects, we introduce an evaluation protocol that complements global metrics with gap- and dispersion-based measures. Our experiments show that HBR significantly reduces recall gaps—by up to 32\% on CIFAR-10 and 16\% on CIFAR-100, outperforming standard frequency-based resampling. We further show that we can improve fairness outcomes by selectively using the hardest samples from a state-of-the-art diffusion model, rather than randomly selecting them. These findings demonstrate that data balance alone is insufficient to mitigate class bias, necessitating a shift toward hardness-aware approaches.}


\maketitle

\vfill
\begin{center}
\small This paper is under review at Springer's \textit{Machine Learning} journal.
\end{center}
\vfill

\section{Introduction}

Class bias, the phenomenon where machine learning classifiers exhibit disparate performance across categories, remains a fundamental challenge in fairness-aware learning and safety-critical applications \citep{zhu2026reducing, li2025bias, katare2025analyzing, brzezinski2024properties, rahman2013addressing, khushi2021comparative, zong2022medfair, tasci2022bias}. To date, this bias has been studied almost exclusively through the lens of data imbalance \citep{hashimoto2018fairness}. This focus is conceptually intuitive: the standard empirical risk minimization (ERM) objective assigns equal weight to all samples, allowing majority ('head') classes to dominate optimization dynamics at the expense of minority ('tail') classes \citep{hashimoto2018fairness}. Consequently, despite data imbalance not being the sole factor contributing to class bias, it has become the most convenient and well-understood stand-in for the problem. This reliance on frequency-centric approach has created a significant research gap: the persistence of class bias in datasets where class frequencies are perfectly balanced.

\begin{figure}[b!]
    \centering
    \includegraphics[width=\textwidth]{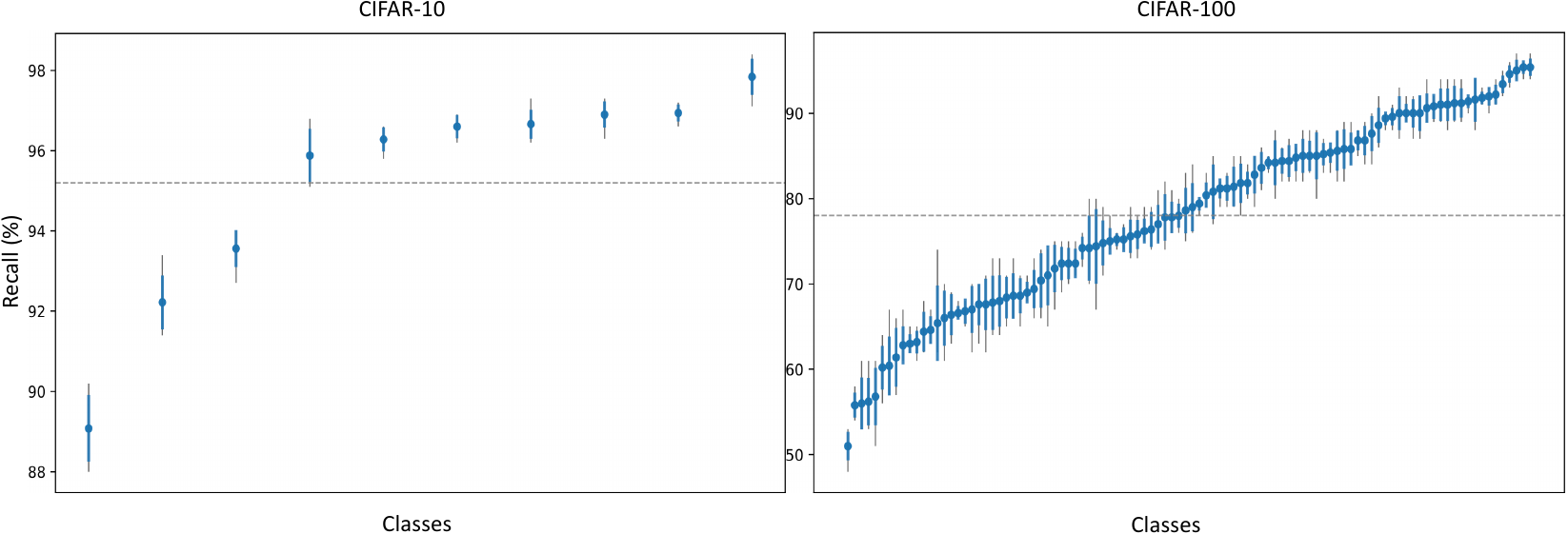}
    \caption{Training five ResNet18 networks on CIFAR-10 and CIFAR-100 reveals large recall gaps across classes, despite the data-balanced nature of these datasets. This highlights that classes exhibit varying degrees of inherent difficulty, a phenomenon we term \textit{hardness imbalance}. This observation motivates Hardness-Based Resampling (HBR), which allows addressing class bias in data-balanced settings.}
    \label{fig:fig1}
\end{figure}

We argue that the prevailing assumptions underlying reweighting and resampling, that data-balanced datasets are optimal \citep{rahman2013addressing} and that 'tail' classes are inherently more difficult solely due to under-representation, do not hold universally. Empirical evidence suggests that class difficulty is often decoupled from frequency \citep{sinha2020class, sinha2022class}. Our experiments with ResNet18 architectures \citep{he2016deep} on CIFAR-10 and CIFAR-100 \citep{krizhevsky2009learning} highlight this disconnect (see Fig. \ref{fig:fig1}). Despite perfect data balance, we observe profound performance gaps: on CIFAR-10, recall for the 'cat' class lags at 89\% compared to a 95\% global average. In CIFAR-100, the disparity is even more acute, with 'motorcycle' recall (95\%) nearly doubling that of the 'boy' class (51\%). These disparities cannot be explained by frequency-based differences, revealing that class bias is far more nuanced than commonly assumed, with data imbalance being merely one of the contributing factors.

We propose a shift in perspective: addressing class bias requires moving beyond frequency to analyze the unequal distribution of difficulty, or hardness imbalance. While hardness is a recognized heuristic in curriculum learning, data pruning, and many other fields \citep{wang2021survey, soviany2022curriculum, settles2009active, ren2021survey, toneva2018empirical, paul2021deep, sorscher2022beyond, agarwal2022estimating, yang2024generalized, pleiss2020identifying, maini2022characterizing, jia2023learning}, it has been overlooked as a primary driver of class-level bias. Existing hardness-aware reweighing and resampling methods, such as Focal Loss \citep{lin2017focal} or ADASYN \citep{he2008adasyn}, are almost exclusively deployed in imbalanced settings to compensate for low sample counts. Even when reweighting is applied to balanced data, it is typically used to improve training efficiency rather than to mitigate systematic performance gaps. By treating hardness as a primary concern rather than a secondary effect of frequency, we identify a 'blind spot' in current mitigation strategies: the possibility of mitigating bias by addressing the intrinsic complexity of specific classes.

To bridge this gap, we introduce Hardness-Based Resampling (HBR), which leverages instance-level hardness estimates to guide resampling. To evaluate its impact on class bias, we introduce a series of gap- and dispersion-based metrics alongside traditional global metrics. Our results show that HBR consistently reduces recall-based class bias. Crucially, the mechanism differs from data-imbalanced settings: there, random oversampling reduces bias simply by balancing frequencies; here, with frequencies already balanced, only oversampling that adds new and informative content can help. Accordingly, random oversampling and SMOTE yield negligible changes, whereas a state-of-the-art diffusion model \citep{karras2022elucidating} produces substantial reductions—closing the recall gap between average easy and hard classes by $0.014$ on CIFAR-10 (from $0.043$ to $0.029$) and $0.027$ on CIFAR-100 (from $0.165$ to $0.138$). We further find that when generative models are involved, harder synthetic samples are more beneficial: selecting samples based on hardness rather than chance nearly doubles the reduction in class bias. This suggests that improving generative models' ability to produce hard samples—rather than the easy samples, which they currently favor \citep{wang2026difficulty}—could yield further gains. Importantly, these reductions do not come at the expense of overall accuracy, which in most cases increases slightly. Altogether, these findings demonstrate that \textbf{class bias is not merely a symptom of data imbalance, but a more fundamental phenomenon rooted in the unequal distribution of hardness across classes}. If we are to build models that are truly fair across all classes, regardless of their frequency, the field must pivot from frequency-centric corrections toward hardness-aware methods that address bias where it actually originates.

Our work provides the following contributions:
\begin{itemize}
    \item We show that hardness imbalance is a critical source of class bias that persists even when dataset sizes are equal. We demonstrate that traditional frequency-based balance is a \textit{necessary but insufficient} condition for equitable class-wise performance.
    \item We empirically demonstrate that hardness-aware resampling significantly reduces recall disparities in balanced datasets. Our analysis identifies two primary factors: the degree of induced imbalance and the quality of oversampled data.
    \item We show that selectively leveraging the hardest samples generated by a diffusion model reduces recall gaps far more than using random generative samples. This suggests that further gains can be made by improving generative models' ability to produce hard samples, rather than the easy samples which they currently favor.
    \item We establish \textit{novel measures of class bias} that integrate global performance metrics with gap- and dispersion-based measures. These measures provide a more granular view of class bias that standard macro-averaging fails to capture.
\end{itemize}

\section{Background}
\label{sec2}

\subsection{Survey on data and hardness imbalances}

\textbf{Data imbalances.} Data imbalance is a phenomenon that only relates to the density of data sampling and can occur both between and within classes. Between-class data imbalance is the most well known and studied problem \citep{leevy2018survey, kaur2019systematic}. It is researched both in binary setting (minority vs. majority class) as well as in the multiclass scenario, in which case the class cardinalities often follow Pareto distribution with few head classes that have high cardinality and many tail classes with low cardinality. 

Data imbalance can also occur within classes, as was first observed by \citet{holte1989concept}. This data imbalance is characterized by nonuniform sampling from class manifolds, resulting in data clusters (disjuncts) of various sizes within each class, with small disjuncts being harder to learn. \citet{japkowicz2001concept} reported that, at the time, \textit{no works took into consideration the fact that both between-class and within-class imbalances may occur}. Later, \citet{jo2004class} made a claim that focusing solely on between-class data imbalance will not always improve the performance, arguing that the core issue behind class bias is the existence of small disjuncts which, as they argue, occur as the consequence of between-class data imbalance. In other words, they claimed that between-class data imbalance isn't a problem in itself, but rather that it causes the emergence of within-class data imbalance which is the main issue causing degradation in classifiers' performance. Despite that, multiple works report lack of standardization \citep{guo2008class} and research \citep{leevy2018survey, ren2023systematic} on within-class data imbalance, and the exact importance of this problem remains an open question.

\textbf{Hardness imbalance.}
Hardness-based imbalance shifts the analytical focus from data cardinalities to model-driven difficulty. Although rarely formalized in existing literature, this phenomenon is a natural corollary of hardness-aware methodologies that utilize heuristic signals to quantify learning difficulty \citep{zhu2024exploring, seedat2024dissecting}. Traditionally, these signals serve as the foundation for identifying problematic samples \citep{pleiss2020identifying, agarwal2022estimating, yang2024generalized}, guiding data pruning \citep{toneva2018empirical, paul2021deep}, or dynamically adjusting training algorithms \citep{lin2017focal, soviany2022curriculum, yuan2025instance}.

We define hardness imbalance as the non-uniform distribution of learning difficulty across objects of interest, ranging from individual samples to entire classes. Within this framework, data imbalance is considered as one of the factors affecting class hardness \citep{lorena2018data, lorena2019complex}. While a rigorous, model-independent definition of "hardness" remains an open theoretical challenge, its effects are empirically accessible. We observe these manifestations through two distinct channels: performance-based metrics (e.g., accuracy) \citep{sinha2020class} and hardness estimates (e.g., confidence, margin, or forgetting events) \citep{toneva2018empirical, pleiss2020identifying}. By focusing on these observable signals, we can reason about the disparate impacts of difficulty without requiring a closed-form analytical definition.

\begin{figure}[t!]
    \centering
    \includegraphics[width=\textwidth]{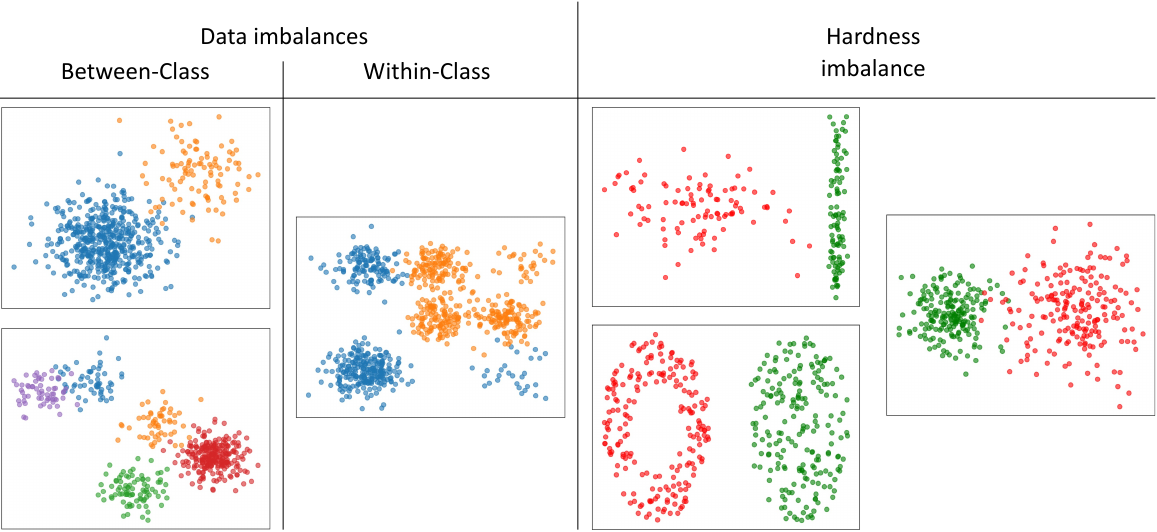}
    \caption{Data imbalance focuses solely on cardinality, be it between classes or within them. On the other hand, hardness imbalance assumes that some classes are harder to learn than others due to geometrical, topological, and sampling-related factors. It provides an explanation into the emergence of class bias in data-balanced datasets.}
    \label{fig:fig2}
\end{figure}

The primary distinction between data and hardness imbalances lies in their underlying assumptions. Data imbalance attributes class bias solely to differences in sampling density. In contrast, hardness imbalance provides a more comprehensive explanation where sampling-related issues are merely one of several contributing factors to poor performance. Hence, hardness imbalance can persist in perfectly data-balanced settings, whereas data imbalance is by definition eliminated once sampling densities are equalized. Although this perspective introduces greater complexity—requiring a consideration of the interplay between geometric, topological, and sampling-related factors—it provides a more general framework for reasoning about the emergence of class bias. We visualize these imbalances in Figure~\ref{fig:fig2}.

\subsection{Hardness estimators}

Hardness estimators can be divided into data- and model-based ones. The former use geometrical or topological properties of class manifolds, or their latent representations, to estimate hardness. Research showed that classes with higher intrinsic dimension \citep{ansuini2019intrinsic, pope2021intrinsic, ma2024unveiling}, Lebesgue measure \citep{ma2023delving}, and curvature \citep{kienitz2022effect, kaufman2023data, ma2023curvature} are harder to learn for models resulting in lower accuracy. This means that incorporating geometry-aware regularization techniques can decrease the class bias and improve the overall performance \citep{ma2023curvature, ma2023delving, ma2024unveiling}. However, applying these methods to input space is problematic due to the curse of dimensionality. This, paired with their higher computational complexity, is why data-based estimators are not popular in curriculum learning, active learning, data pruning, and other fields of machine learning using hardness-aware methods. In those fields model-based estimators reign supreme.

Confidence-based estimators are by far the most popular in the literature due to their simplicity. They are defined as:
\begin{align}
    C = p_{\hat{y}},
\end{align}
where $p_{\hat{y}}$ is the probability of the true class $\hat{y}$. The higher this value, the more confident the model is in its prediction indicating how easy the sample is for it to learn. Conversely, hard samples are the ones that model is not confident in. This estimator was used by \citet{lin2017focal} in their Focal Loss. 

\begin{figure}[t!]
    \centering
    \includegraphics[width=\textwidth]{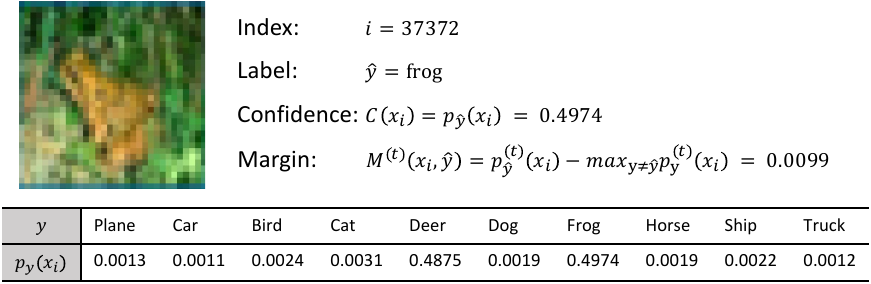}
    \caption{Example of hardness estimation using confidence and margin on an image from CIFAR-10 using a trained ResNet-18 model. Confidence suggest image of medium hardness, while margin correctly identifies the confusion between a deer and a frog.}
    \label{fig:fig3}
\end{figure}

Confidence-based estimators rely solely on a single logit, without taking uncertainty into consideration. This issue is partially addressed by margin-based estimators which consider the difference between the correct class logit and the strongest competing class \citep{tao2023dual}. More formally, the margin is defined as:
\begin{align}
    M^{(t)}(x, \hat{y}) = z^{(t)}_{\hat{y}}(x) - max_{i \neq \hat{y}} z^{(t)}_i(x),
\end{align}
where $t$ is the training epoch at which the computation was performed. The larger the margin the easier the sample is considered to be. Meanwhile, low and negative margins indicate that model was not certain regarding its prediction or that it made a mistake (see Figure \ref{fig:fig3} for example). To increase the robustness of estimators the computation of margin can be extended to involve learning dynamics. An example of estimator doing so is Area Under Margin (\textbf{AUM}) which computes margin for each sample after every training epoch \citep{pleiss2020identifying}. Formally, :
\begin{align}
    \text{AUM}(x, \hat{y}) = \frac{1}{T}\sum_{t=1}^TM^{(t)}(x, \hat{y}).
    \label{eq: AUM}
\end{align}

Learning dynamics contain a wealth of information and their use is very popular when estimating hardness. One of the core properties of data sample gathered through analyzing learning dynamics is forgetting \citep{toneva2018empirical}.Forgetting event occurs when a model transitions from correctly classifying a sample at epoch $t$ to misclassifying it at epoch $t+1$. Research shows that the frequency of forgetting events is not uniformly distributed across samples: some are frequently forgotten throughout training, while others are consistently remembered, with some never being forgotten. Hence, hardness of a data sample is proportional to the number of its forgetting events—the more often it is forgotten by the model during training the harder it is. Crucially, confidence, margins, and forgetting statistics are often highly consistent across different model architectures highlighting that while forgetting is a model-based estimates it is able to capture the hardness inherent in data.

To summarize, while hardness in itself is an intrinsic property of data, estimating it is very challenging in practice. Therefore, using model-based heuristics is much more popular and is also an approach that we will take in this work. Specifically, we will use AUM \citep{pleiss2020identifying} which was shown by \citet{seedat2024dissecting}, the only systematic benchmarking work on hardness estimation, to produce one of the most accurate and stable estimates.

\subsection{Oversampling and resampling methods} \label{sec2.3}

\textbf{Undersampling.} Hardness-based pruning has gained recognition for its ability to identify and remove data samples with minimal impact on model performance \citep{toneva2018empirical, paul2021deep, maini2022characterizing}. \citet{sorscher2022beyond} provide particularly compelling evidence of its effectiveness, showing that this approach can break beyond power-law scaling of error with respect to dataset size. Since data pruning has the same purpose as undersampling—removing specified number of samples with minimal negative impact on performance—we also use this approach in our work.

\textbf{Oversampling.} Oversampling can take several forms. One can simply replicate existing data (random oversampling), apply data augmentation to create synthetic variants \citep{ahn2023cuda, vu2024lcsl}, or interpolate between minority class samples (e.g., SMOTE \citep{chawla2002smote} and its variants such as Borderline-SMOTE \citep{han2005borderline}, ADASYN\citep{he2008adasyn}, or DeepSMOTE \citep{dablain2022deepsmote}). However, due to the usage of duplication or interpolation these methods typically do not generate genuinely new examples and may lead to overfitting. To overcome this limitation, recent works have explored generative models (e.g., Generative Adversarial Networks \citep{goodfellow2020generative}, or diffusion models \citep{ho2020denoising}) as a way to synthesize novel samples that better capture the true class distribution. These generative approaches have increasingly become a focal point for addressing data scarcity \citep{ali2019mfc, zheng2020conditional, marchesi2023generative}. Unfortunately most, if not all, resampling methods have been evaluated exclusively on imbalanced datasets.

\textbf{Generative models and their evaluation} Generative models aim to produce novel synthetic samples, often with the goal of filling gaps in the data manifold caused by non-uniform sampling distributions. Their architectures have developed over the years, moving from Variational Autoencoders \citep{doersch2016tutorial} and Generative Adversarial Networks \citep{goodfellow2020generative} to Diffusion Models \citep{ho2020denoising}, which are the current state of the art in this field. Considering their purpose, it is imperative to use metrics that measure the quality of the generated samples. The two most commonly used ones are the Inception Score (IS) \citep{salimans2016improved} and the Fréchet Inception Distance (FID) \citep{heusel2017gans}. The former is a proxy-based metric that measures the entropy of a batch of synthetic images. A higher IS is interpreted as higher-quality synthetic samples and indicates that the Inception V3 model \citep{szegedy2016rethinking}, used as the proxy, was able to correctly and confidently classify the majority of the synthetic samples. The latter works by comparing the distribution of latent representations produced by Inception V3 for real images from a reference set and for the synthetic samples, computing the squared Wasserstein distance between the two multivariate Gaussian distributions. A lower FID means that the distribution of the synthetic samples is very similar to that of the real data.

Despite their popularity, these metrics exhibit important shortcomings. IS implicitly favors generative models that produce samples which the Inception network can classify with high confidence—typically the easier regions of the data manifold. As a result, models optimized for IS may underrepresent rare or ambiguous cases that are crucial for downstream robustness. FID, while conceptually broader, also has fundamental limitations. It measures only global alignment between real and synthetic feature distributions, assuming they can each be represented as a single multivariate Gaussian. This assumption collapses the complex structure of data manifolds, erasing distinctions between classes and overlooking whether harder regions are faithfully captured. Moreover, because FID depends on representations learned by a specific pretrained model, it inherits that model’s inductive biases. In practice, this suggests that FID can overestimate the fidelity of generative models whose samples resemble familiar, low-hardness regions, while failing to penalize insufficient coverage of harder ones.

These concerns lead to tangible issue. Namely, as demonstrated by \citet{wang2026difficulty}, the hardness spectrum of synthetic data is often skewed toward easier samples compared to real data. While a small number of specialized architectures have attempted to mitigate this by explicitly encouraging the generation of harder samples \citep{vandenhende2019three, pennisi2021self, ferracci2024targeted, wang2026difficulty}, research in this direction remains remarkably limited. Consequently, the relationship between synthetic sample hardness and downstream model performance is not yet well understood. It remains unclear whether harder synthetic samples are inherently more valuable for generalization, whether the optimal synthetic hardness should strictly mirror that of real data, or whether pushing models toward high-hardness regions risks producing noisy or unrealistic outputs. This work aims to partially address the first of these open problems.

\section{Methodology}
\label{sec3}

\subsection{Formal Experimental Design}

\begin{figure}[t!]
    \centering
    \includegraphics[width=\textwidth]{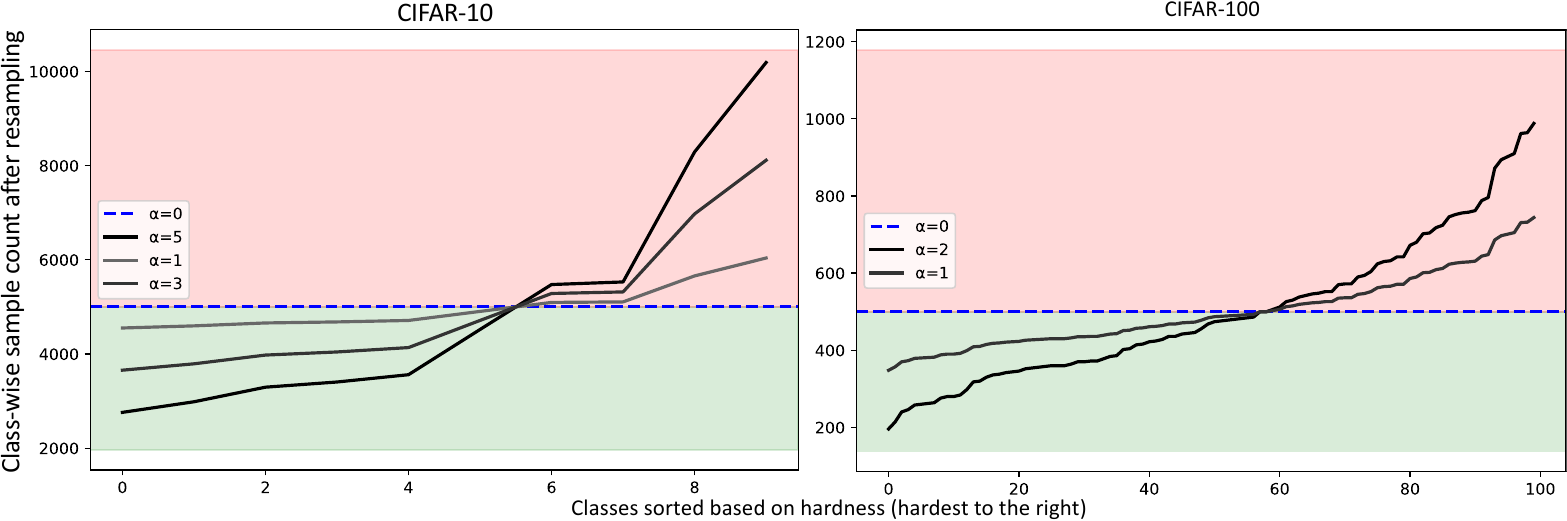}
    \caption{Visualization of the degree of the data imbalance introduced through HBR. Adjusting the $\alpha$ parameter allows us to have a finer control over the severity of the introduced imbalance.}
    \label{fig:fig4}
\end{figure}

We define Hardness-Based Resampling (HBR) as a data preprocessing operation that breaks the data balance by resampling classes based on their hardness, changing the number of samples in class $c$ to some $b_c$ that is proportional to the true hardness of class $c$, $\hat{h}_c$. It is composed of two steps: 1) undersampling; and 2) oversampling. Undersampling works by removing the $a_c - b_c$ easiest samples from classes where $b_c < a_c$, where $a_c$ is the cardinality of class $c$ before resampling. Meanwhile, oversampling is performed by expanding the training set by $b_c - a_c$ samples. In balanced setting, which we consider in this work, $a_i = a_j$ for all $(i, j) \in \{1, ..., k\}^2$. Naturally, ground-truth hardness $\hat{h}_c$ is inaccessible, forcing us to rely on hardness estimates. Specifically, we use AUM to obtain instance-level hardness estimates $h_i$, which we later convert to class-level ones through averaging:
\begin{equation}
    H_c = \frac{\sum_{i=1}^n h_i \times  \mathbf{1}_{\{y_i=c\}}}{\sum_{i=1}^n \mathbf{1}_{\{y_i=c\}}},
    \label{eq:equation2}
\end{equation}
where $\mathbf{1}_{\{y_i=c\}}$ is an indicator function that equals $1$ if $y_1 = c$ and $0$ otherwise, and $n$ is dataset's cardinality. Intuitively, the resampling ratios $b_c$ should be based on those class-level estimates as follows:
\begin{align}
    b_c = \left\lfloor f \left( \frac{{H}_c}{\sum_{c'} H_{c'}} \sum_c a_c  \right) \right\rfloor,
    \label{equation 4.1}
\end{align}
with $f: \mathbb{R}{+} \rightarrow \mathbb{R}{+}$ being a function that determines the rate of progression. For instance, an identity function yields linear progression, while logarithmic function puts more emphasis on easy classes than hard ones, as it amplifies the differences in hardness across classes more significantly for the easiest classes. Since Equation \ref{equation 4.1} does not allow for modifying the degree of the introduced data imbalance we update the number of samples after resampling as follows:

\begin{align}
    b'_c = a_c + \lfloor \alpha \left( b_c - a_c \right) \rfloor
    \label{equation 4.2}
\end{align}
with $b'_c$ being the updated sample count for class $c$ and $\alpha$ a scalar that controls the degree of the introduced data imbalance. In Figure \ref{fig:fig4} we show how different $\alpha$ values used in our experiments affect the degree of the introduced data imbalance.

\subsubsection{Designing case studies.} To reiterate, we investigate whether the fairness objective can be met by performing HBR. This objective has the following key challenges: 1) oversampling techniques struggle with generating genuinely new, i.i.d. data samples; and 2) estimating resampling ratio relies on class-level hardness estimates, the quality of which is hard to ascertain. While the second point is very challenging, and ultimately remains a limitation of HBR approach, or any hardness-aware method for that matter, the first one can be addressed through a carefully designed case study. We therefore design two complementary case studies: Case Study 1, which introduces a holdout set to alleviate the questionable quality of synthetically generated samples, and Case Study 2, which more closely reflects realistic training conditions, where no access to holdout set is provided. 

\textbf{Case Study 1: Access to holdout set.} We assume that CIFAR-10 and CIFAR-100 were obtained through i.i.d. sampling from an underlying distribution $\mathcal{X}$. More specifically, for each class $c$, $\mathcal{D}_c$ is a set of i.i.d. samples drawn from $\mathcal{X}_c$, which is a distribution restricted to class $c$. Now consider a sub-dataset $\mathcal{D}' = \bigcup_{c=1}^C \mathcal{D}_c'$, where $\mathcal{D}_c' \subset \mathcal{D}_c$ and $|\mathcal{D}_c'| = a'$ for each class $c$, with $a' < a$. Next, define a holdout set $\mathcal{D}_c^{\text{ho}} := \mathcal{D}_c \setminus \mathcal{D}_c'$. We assume that $\mathcal{D}_c' \sim \mathcal{X}_c$ and that $\mathcal{D}_c^{\text{ho}} \sim \mathcal{X}_c$, making $\mathcal{D}_c^{\text{ho}}$ a simulated oracle for new draws from $\mathcal{X}$. The $\mathcal{D}_c^{\text{ho}} \sim \mathcal{X}_c$ assumption allows us to alleviate issue (1). We call this hypothetical situation in which the dataset construction process was stopped early \textit{Scenario A}. In order to simulate it we construct $\mathcal{D}_c'$ by randomly selecting $a'$ samples from each class without replacement. The remaining samples form a disjoint holdout set $\mathcal{D}^{\text{ho}}$. Although subsampling without replacement technically induces weak dependence between samples, we treat the resulting subsets $\mathcal{D}_c'$ and $\mathcal{D}_c^{\text{ho}}$ as approximately i.i.d. in practice. Ultimately, these assumptions play a formal role and do not impact the validity of our conclusions, especially given that CIFAR-10 and CIFAR-100 are themselves not strictly i.i.d. in any meaningful sense. Hence, in this case study we apply HBR to $\mathcal{D'}$, which we will call $\mathcal{D}^{CLP}$ as it was obtained through class-level pruning.

\textbf{Case Study 2: No access to holdout set.} Here we apply HBR to the original dataset $\mathcal{D}$. Hence, the only difference from Case Study 1 is that we do not perform CLP before applying resampling.

\subsubsection{Hardness-Based Resampling}

For undersampling we ranked samples within each easy class by AUM, removing the easiest ones until the desired sample count was achieved. For oversampling we employ three strategies: (1) random oversampling, (2) SMOTE, and (3) synthetic samples generated using EDM \citep{karras2022elucidating}. Since training a diffusion model is computationally prohibitive, we use a million of synthetic samples made publicly available by \citet{wang2023better}. For Case Study 1, we use the holdout set instead of samples generated using EDM, as the latter would require retraining the model on $\mathcal{D}^{CLP}$ to avoid information leakage, which is computationally prohibitive. To address the open question described in Section \ref{sec2.3} regarding the preferential generation of easy samples by generative models we further define three EDM sub-strategies:
\begin{enumerate}
    \item rEDM, where synthetic samples are drawn randomly from the specified classes
    \item aEDM, where samples are drawn from the set of $2b_c$ synthetic images whose hardness is closest to the class average (based on AUM)
    \item hEDM, where samples are drawn from the set of $2b_c$ hardest synthetic images of class $c$
\end{enumerate}

\subsection{Measuring degree of class bias}
\label{sec3.2}

Let's define a balanced dataset $\mathcal{D}$ containing $a$ samples in each class. We are interested in addressing class bias—a phenomenon characterized by inequalities in recall observed across classes. More formally, let $r_c := \text{Rec}(M, \mathcal{D}_c^{\text{test}})$ denote the recall of model $M$ on a test set containing only samples from class $c$. Class bias occurs when some class $c$ performs better or worse than the mean recall across all classes, $\bar{r}_c$. %

To quantify the degree of class bias, we employ a set of metrics that capture complementary aspects of per-class recall disparities:  (i) \textit{gap-based} metrics, which compare the easiest and hardest classes, and (ii) \textit{dispersion-based} metrics, which measure variability across all classes.

\begin{itemize}
    \item \textbf{Gap-based metrics:}
    \begin{itemize}
        \item \textbf{Maximum gap:}
        \begin{equation}
            \Delta_{m} = \max_c r_c - \min_c r_c,
        \end{equation}
        measures the gap between the highest and lowest class recalls. It captures extreme inequality, making it highly sensitive to outliers.

        \item \textbf{Quantile gap:}
        \begin{equation}
            \Delta_{q} = \frac{1}{k}\sum_{i=1}^{k} r_{(K+1-i)} - \frac{1}{k}\sum_{i=1}^{k} r_{(i)},
        \end{equation}
        where \(r_{(i)}\) denotes the \(i\)-th smallest element of \(\mathbf{r} = (r_1, \dots, r_C)\). 
        This measures the average gap between the \(k\) easiest and \(k\) hardest classes. 
        By averaging across multiple classes, it reduces sensitivity to single outlier classes while still emphasizing extremes like \(\Delta_m\). 
        We set \(k=2\) for CIFAR-10 and \(k=10\) for CIFAR-100.

        \item \textbf{Hardness-based equality gap:}
        \begin{equation}
            \Delta_{he} = \frac{1}{|\mathcal{E}|}\sum_{c \in \mathcal{E}} r_c - \frac{1}{|\mathcal{H}|}\sum_{c \in \mathcal{H}} r_c,
        \end{equation}
        where \(\mathcal{E}\) and \(\mathcal{H}\) denote sets of easy and hard classes identified through HBR ratios. 
        This measure directly reflects our experimental design by comparing the mean performance of classes targeted for under- vs.\ oversampling, making it more robust to individual outlier classes than \(\Delta_m\) or \(\Delta_q\).
    \end{itemize}

    \vspace{3pt}

    \item \textbf{Dispersion-based metrics:}
    \begin{itemize}
        \item \textbf{Standard deviation:}
        \begin{equation}
            \Delta_{\sigma} = \text{Std}(\mathbf{r}),
        \end{equation}
        quantifies the overall dispersion of class recalls around their mean. It provides alternative way to measure the degree of class bias, accounting for all classes simultaneously. However, since it squares deviations, it tends to overweight large differences in skewed distributions, making it more sensitive to outliers than \(\Delta_{he}\).

        \item \textbf{Median absolute deviation:}
        \begin{equation}
            \Delta_{\text{MAD}} = \text{Median}\bigl(|r_c - \text{Median}(\mathbf{r})|\bigr),
        \end{equation}
        is a robust alternative to \(\Delta_\sigma\) that downweights classes with extreme recall values. It offers a stable measure of the degree of class bias even in the presence of outlier classes.
    \end{itemize}
\end{itemize}

While $\Delta_m$ and $\Delta_q$ are more sensitive to outliers, comparing them to other metrics reveals whether changes to class bias are localized to a few extreme classes or spread more uniformly across all classes. In addition, we track the average performance (Recall and Precision) to contextualize the trade-off between the changes in class bias and accuracy. Our objective is therefore to minimize the $\Delta$ metrics while maintaining the highest possible class-averaged performance. 

\subsection{Dataset description and experimental setup}
\label{sec: data_description}

CIFAR-10 contains $60,000$ 32x32px color images in 10 classes, with $6,000$ images per class. The CIFAR-100 contains $60,000$ 32x32px color images in $100$ classes, with $600$ images per class. Following the standard PyTorch \cite{paszke2019pytorch} partitioning, both datasets are split into training and test sets of sizes $50,000$ and $10,000$, respectively, with uniform class distribution. 

In our main experiments, we train ensembles of ResNet-18, modified for low-resolution data, for $200$ epochs using SGD (lr $0.1$, momentum $0.9$, weight decay $0.0005$), with a $0.2$ learning rate decay at epochs $60$, $120$, and $160$, and a batch size of $128$. Further information regarding more detailed experimental design, ensuring statistical reliability and reproducibility, and design of the paired t-test are available in Appendices \ref{secA1}, \ref{secA2}, and \ref{secA3}, respectively.

Our choice of these datasets and the ResNet architecture is a deliberate one, aligned with established practices in hardness-aware literature \citep{toneva2018empirical, pleiss2020identifying, paul2021deep, maini2022characterizing, mindermann2022prioritized, kaufman2023data}. These environments are complex enough for hardness imbalance to significantly impact fairness, yet small enough to allow for the extensive experimental iterations and multi-seed analysis required for statistical rigor. Because hardness is an inherent data property, its fundamental effects are observable regardless of dataset scale; we therefore prioritize the granular and reproducible characterization enabled by this setup over the computationally prohibitive breadth of larger-scale benchmarks.

\section{Results}
\label{sec4}
\begin{figure}[t!]
    \includegraphics[width=\textwidth]{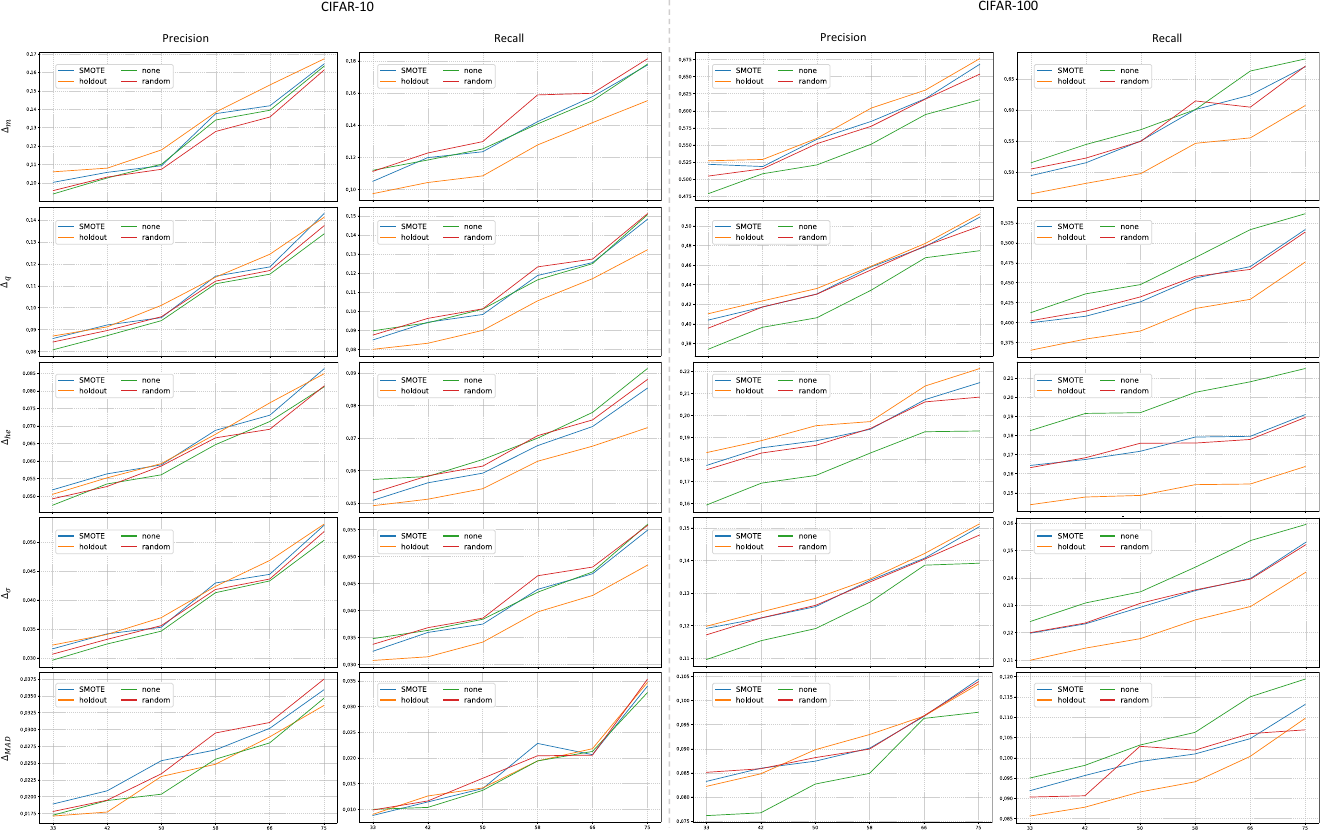}
    \caption{\textbf{Case Study 1:} Analyzing changes in various fairness metrics on pruned versions of CIFAR-10 and CIFAR-100 reveals that HBR reduces the recall gap across classes at the cost of increasing the precision gap (here the lower the $\Delta$ metrics the lower the performance gaps across classes). We find that HBR yields similar improvements no matter the pruning rate. Furthermore, it seems to be highly reliant on the quality of samples used for oversampling, as random- and SMOTE-based resampling was not able to meaningfully impact the performance gap. The changes to class bias are more pronounced on CIFAR-100, most likely due to more linear-like hardness spectrum of this dataset (see Figure \ref{fig:fig1}). We report means over four ensembles of four models trained on 4 distinct $\mathcal{D}^{CLP}$ and do not report standard deviation for clarity.}
    \label{fig:fig5}
\end{figure}

In this section, we analyze the results of both case studies and propose possible explanations for the observed trends. Figure \ref{fig:fig5} presents the changes in the fairness metrics, defined in Section \ref{sec3.2}, for the first case study on CIFAR-10 and CIFAR-100, while Figure \ref{fig:fig6} shows the corresponding results for the second case study. In Appendix \ref{secA4}, we report the outcomes of the Student’s t-test on both case studies and datasets and include further analysis. 

\begin{figure}[t!]
    \includegraphics[width=\textwidth]{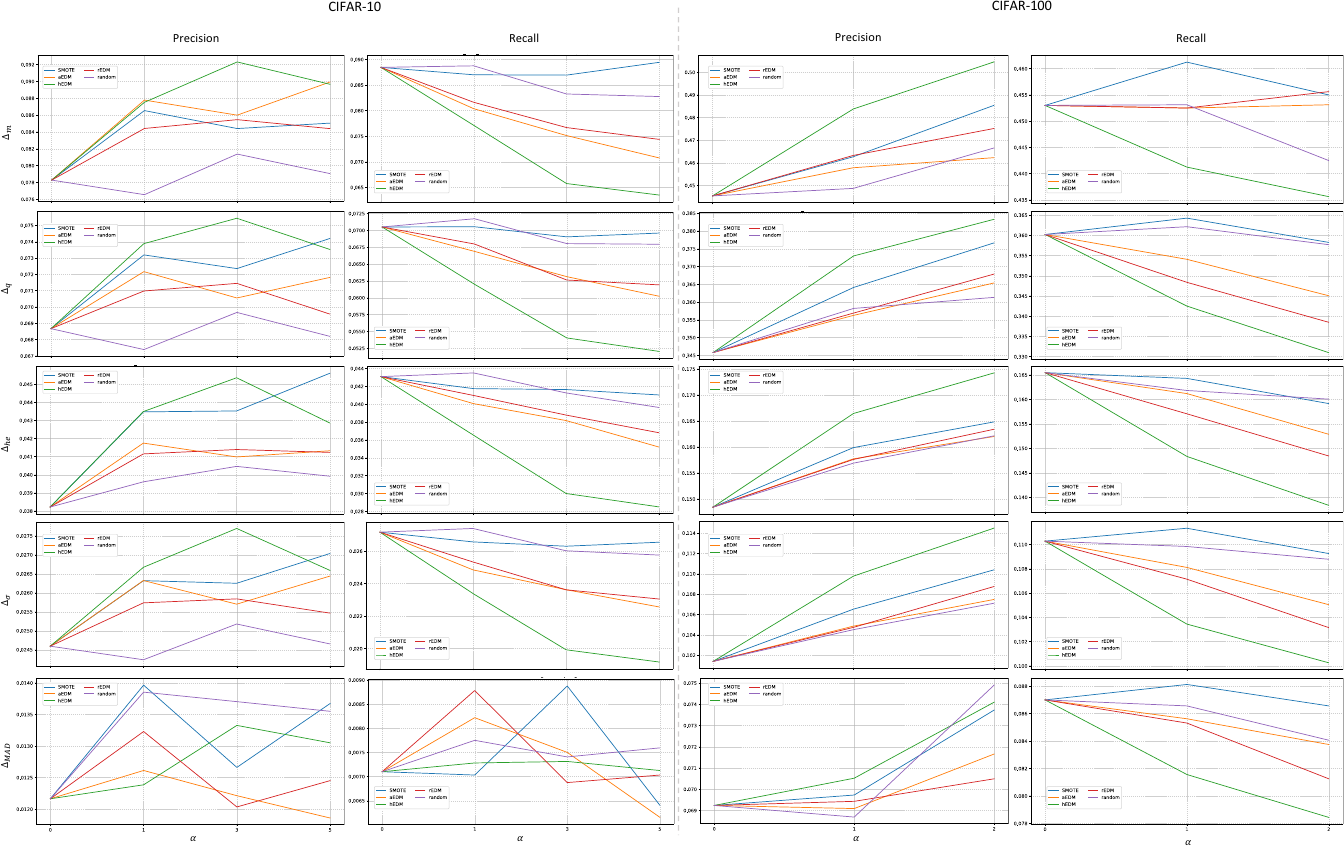}
    \caption{\textbf{Case Study 2:} We notice that, in most cases, the higher the degree of the introduced data imbalance (x-axis) the larger the effect on fairness when high quality synthetic samples are available. We expect the existence of some threshold $\alpha$ beyond which the drawback from too heavy data imbalance starts to overwhelm the benefits of addressing hardness imbalance. Unlike in the first case study (Figure \ref{fig:fig5}, here we see hardness-based imbalance having similar effects on both datasets. We attribute this to the subpar quality of EDM-generated samples compared to the real samples from the holdout set. This is evidenced by the discrepancy between the forecasted fairness metrics from the first study and the higher values actually observed in this figure (e.g., $\Delta_{he}$ was forecasted to drop below $0.14$ for $\alpha = 1$ on CIFAR-100).}
    \label{fig:fig6}
\end{figure}

\textbf{Sample quality is crucial for HBR.} We first observe a consistent decrease in recall-based class bias resulting from HBR. In the first case study, we find statistically significant decreases in $\Delta_m$, $\Delta_q$, $\Delta_{he}$, and $\Delta_\sigma$ when using the holdout set for oversampling. Specifically, across all pruning rates, HBR decreases the recall gap between the average hard and easy class ($\Delta_{he}$) by approximately $0.01$ on CIFAR-10 and $0.04$ on CIFAR-100. Similar trends appear in the second case study when only the hardest EDM samples are used for oversampling (hEDM): $\Delta_{he}$ decreases from $0.043$ to $0.029$ on CIFAR-10 and from $0.165$ to $0.138$ on CIFAR-100. Using random EDM samples (rEDM) or average EDM samples (aEDM), whose hardness most closely matches the target class hardness, also frequently reduces recall gaps, although less consistently. In contrast, random oversampling and SMOTE generally yield smaller or statistically insignificant changes, and occasionally worsen the class bias. The importance of high quality synthetic samples is particularly evident in Figure \ref{fig:fig5} on CIFAR-100, where we see that changing from SMOTE or random oversampling to holdout set almost doubles the changes in recall-based metrics.

These observations yield two important insights. (i) Unlike standard class imbalance, where simple oversampling succeeds by increasing class cardinality and thus gradient impact, hardness imbalance requires informative samples that populate low-density and/or high-complexity regions—simple replication or interpolation cannot achieve this. (ii) Our results directly answer the earlier open question about hardness in generative modeling: \textit{harder synthetic samples (hEDM) consistently improve generalization more than easy or medium-difficulty ones}, highlighting that steering generative models toward hardness is a promising research direction. Hence, from now on we will focus on results on holdout set or hEDM in our analysis, unless specified otherwise.

\textbf{Quality of EDM-generated samples is insufficient.} Analysing the results across case studies reveals further nuances. On CIFAR-10, the improvements from HBR are remarkably consistent across both case studies: $\Delta_m$ and $\Delta_q$ improve by $\approx 0.01$, while $\Delta_{he}$ and $\Delta_\sigma$ improve by $\approx 0.005$. This means that on CIFAR-10, the hardest EDM-generated images enable bias reductions comparable to those achieved with the holdout set, suggesting they are of sufficient quality for this task. However, this parity disappears on CIFAR-100. While the first case study (holdout set) achieves a $\Delta_{he}$ reduction of $\approx 0.04$, the second case study (hEDM) only achieves $\approx 0.01$. 

We attribute this discrepancy to a shortfall in synthetic informativeness rather than a saturation of the resampling method itself. This is supported by the fact that the bias reduction in the first case study remains scale-invariant across all pruning rates, suggesting that the $\approx 0.04$ improvement is a function of the holdout data's quality and should theoretically translate to the full dataset. Furthermore, direct comparison shows that the holdout set at a $33\%$ pruning rate (Fig. \ref{fig:fig5}) still outperforms hEDM on the full dataset (Fig. \ref{fig:fig6}). These observations suggest that while synthetic hard samples are not yet as informative as their real-world counterparts in more complex datasets, focusing on the "hard" spectrum of generative models is a promising direction that helps narrow this informativeness gap.

\begin{figure}[t!]
    \includegraphics[width=\textwidth]{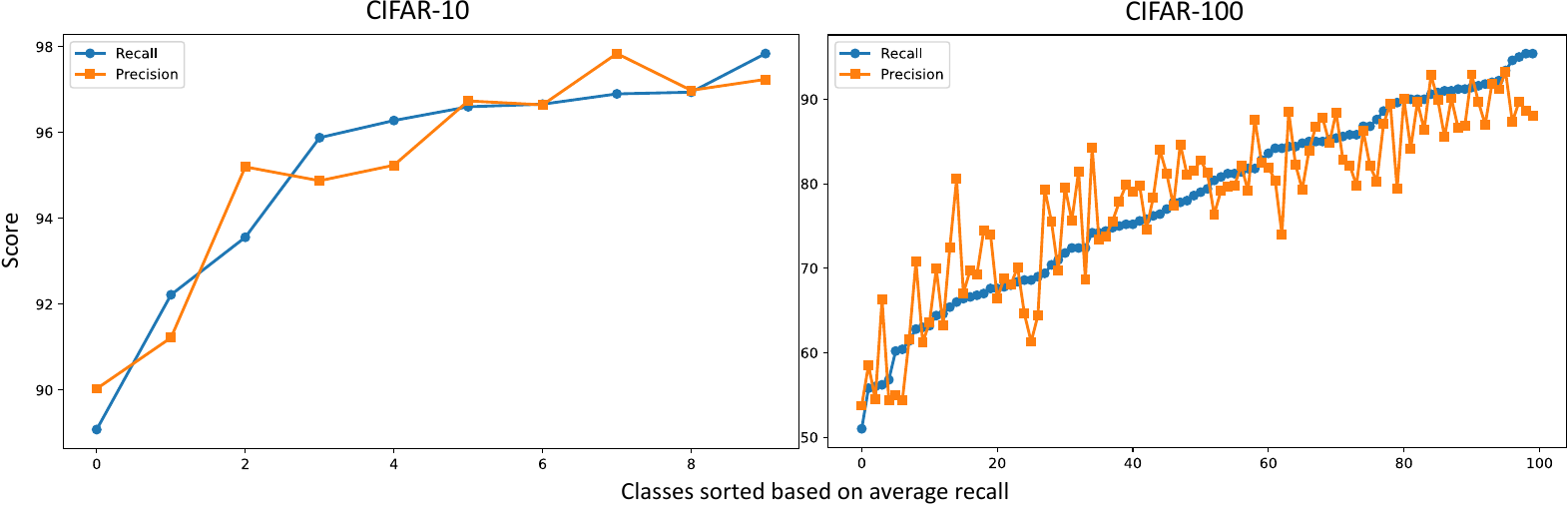}
    \caption{We find that the recall and precision gaps are aligned - in most cases low/high precision of a class is matched with its low/high recall. This indicates that \textit{any attempts in reducing recall gap will lead to increase in precision gap and vice versa} due to the precision-recall trade-off occurring at class level.}
    \label{fig:fig7}
\end{figure}

\textbf{More pronounced data-imbalance leads to higher fairness gains.} Beyond dataset-specific variations, we also analyze how the severity of the induced imbalance (controlled by $\alpha$) modulates these trends (see Figure \ref{fig:fig6}). Consistent with expectations, we observe that in most cases greater imbalance yields stronger and more statistically significant gains from HBR. Although our experiments do not explicitly identify a turning point, we hypothesize the existence of a threshold $\alpha_t$ beyond which the harm from data imbalance outweighs the benefits of mitigating hardness imbalance. Designing an algorithm to efficiently estimate this threshold is beyond the scope of this work. 

\begin{figure}[t!]
    \includegraphics[width=\textwidth]{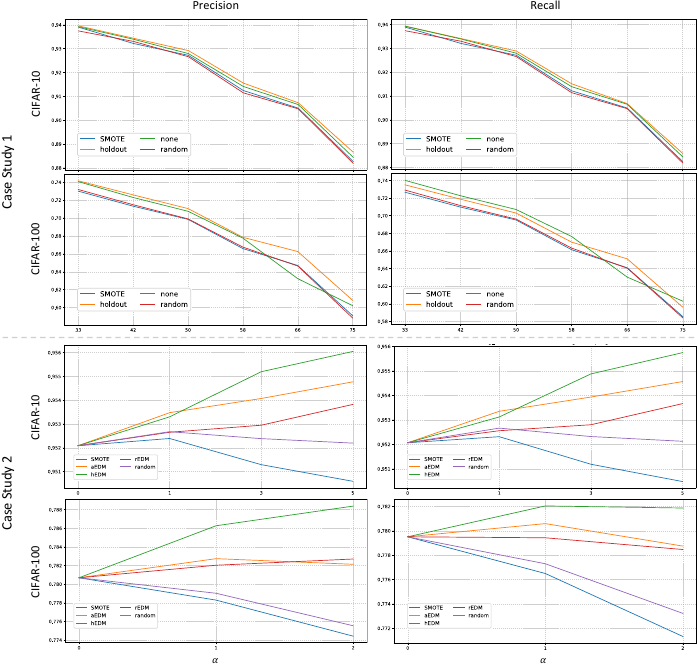}
    \caption{Analysing the changes to average precision and recall we notice that the reduction of class bias, reported in Figures \ref{fig:fig5} and \ref{fig:fig6}, is not accompanied by a significant reduction in overall performance. In fact, we notice that in some cases the average metrics slightly increase (here higher value indicates higher overall performance).}
    \label{fig:fig8}
\end{figure}

\textbf{HBR worsens precision gaps.} Addressing recall-based class bias was the main objective of introducing HBR; however, it is also useful to study the effects that this method has on precision gaps and macro-metrics. We find that the reduction of recall gaps is paired with an increase in precision gaps. As shown in Appendix \ref{secA5}, Figures \ref{fig: CIFAR-10 Tn}–\ref{fig: CIFAR-100 Fp}, oversampling hard classes increases the number of positive predictions for those classes, leading to higher counts of both true positives (TP) and false positives (FP), while false negatives (FN) and true negatives (TN) decrease. This explains the observed increase in recall for hard classes and decrease for easy classes, and the consequential reduction of recall gaps. Naturally, the resulting changes in precision are opposite to those observed for recall. When combined with the observed increase in precision gaps, this suggests that recall and precision gaps are aligned across classes: classes with high recall also tend to have high precision, and vice versa. We verify this empirically on CIFAR-10 and CIFAR-100 (with recall measured as an average over five ResNet18 models), observing strong correlations—Pearson correlations of 0.9395 and 0.9025, and Spearman correlations of 0.9394 and 0.8961 for CIFAR-10 and CIFAR-100, respectively (see Figure \ref{fig:fig7}). This indicates that the precision–recall trade-off cannot be avoided when attempting to reduce class bias. If this correlation were weaker, one could improve both precision and recall gaps simultaneously—for example, by benefiting classes with high recall but low precision. However, under the observed strong alignment, improvements in recall gaps necessarily come at the expense of increased precision gaps.

\textbf{Impact on macro values} A key question is whether these fairness gains come at the expense of overall performance. Examining class-averaged precision and recall, we find that in most situations our method maintains or even slightly improves macro-level metrics across both datasets (see Figure \ref{fig:fig8}, and Tables in Appendix \ref{secA4}). This observation also explains why we focus on precision and recall rather than derived metrics like F1: because both measures consistently move in the same direction—typically increasing together and rarely diverging—a harmonic mean would simply mirror this joint trend without providing additional insight. The observed reductions in recall gaps therefore translate to a net positive or neutral effect on aggregate performance, rather than a tradeoff. Naturally, aggressive resampling with sufficiently large $\alpha$ would eventually degrade performance, but within the range we examine, HBR achieves its fairness goals without compromising accuracy.

\section{Conclusion and future work}
\label{sec5}

In this work we have investigated the relationship between class-wise performance and data distribution, uncovering two important insights into the nature of class bias. First, our empirical evaluation on CIFAR-10 and CIFAR-100 demonstrates that dataset-level balance is an insufficient metric; despite data-balance, these datasets exhibit significant recall disparities across classes. This confirms that aggregate performance metrics frequently mask systematic class bias. Consequently, we advocate for a shift in evaluation measures towards gap- and dispersion-based metrics that can expose these hidden inequities even in perfectly balanced datasets. Second, we have shown that a targeted hardness-based resampling strategy can substantially mitigate these recall gaps while preserving the overall performance. This finding challenges the conventional wisdom that fairness requires a trade-off with performance, suggesting instead that controlled, intentional data imbalance can be a potent tool for counteracting inherent data difficulty. The efficacy of HBR is fundamentally contingent upon the quality of the oversampled data and the magnitude of the induced imbalance. 

Unlike classical data imbalance, for which random oversampling often suffices, hardness imbalance is a more complex phenomenon arising from the intricate geometry and topology of the data distribution. Our empirical findings substantiate this distinction, demonstrating that naive oversampling is insufficient to mitigate hardness-driven disparities.

A compelling frontier for this work lies in the intersection of sample hardness and generative modeling. We observed a significant "hardness vacuum" in contemporary literature, where synthetic samples are disproportionately concentrated in the easier regions of the feature space. Such skew limits the utility of generative augmentation for fairness-oriented tasks, as synthetic data fails to replicate the challenging tail-end of the real distribution. This limitation underscores a \textbf{critical need for hardness-aware generative frameworks} that explicitly incorporate difficulty signals into their training objectives or evaluation protocols. By incentivizing the generation of "informative" rather than merely "high-fidelity" samples, future generative models could become vital tools for mitigating hardness imbalance. 

Ultimately, this research motivates a transition away from the pursuit of static dataset balance and toward an adaptive framework centered on the estimation and correction of hardness-based disparities. We anticipate that refining these adaptive resampling schemes will be a cornerstone in developing models that provide consistent, reliable performance across all classes.

\begin{appendices}

\section{Detailed experimental design}\label{secA1}

\textbf{Hardness estimation.} HBR relies on hardness estimates at two stages: (i) to compute resampling ratios by extension identifying easy and hard classes (Equation \ref{equation 4.1}), and (ii) to guide pruning of easy classes. Ideally, both stages—and both case studies—would rely on identical hardness estimates. However, this is problematic in \textit{Scenario A} (case study 1), where the subsampling alters the relative hardness of many samples: an originally easy sample may become hard simply due to sampling effects. Conversely, recomputing hardness estimates for each scenario would incur significant computational overhead (especially considering the instability of single seed estimates, and cost of producing multiple-seed estimates). We therefore use a single, consistent set of hardness estimates (AUM) across all settings, while acknowledging that this introduces some noise in \textit{Scenario A}.

We compute AUM values by averaging results from ten baseline models trained on unmodified versions of CIFAR-10 and CIFAR-100. We chose this number as we found the estimates to stabilize at this point. We adopt AUM as our default estimator because the benchmark of \citet{seedat2024dissecting} identified it as one of the best-performing hardness estimators. We compute the resampling ratio based on those hardness estimates. We offset each class-hardness by the hardness of the hardest class (the one with lowest AUM) if it is negative to ensure sensible resampling ratios. After that we invert AUM to ensure high values correspond to hard classes (to match equation \ref{equation 4.1}

\textbf{Case Study 1.} 
\textit{Design.} We prune the dataset of interest using thresholds of $33\%$, $42\%$, $50\%$, $58\%$, $66\%$, and $75\%$. Smaller pruning thresholds are omitted because, for these values, the maximum number of samples after resampling ($b_c$) exceeds the available sample count $a_c$ in the holdout set for the hardest classes. After pruning via class-level pruning (\textbf{CLP}), which removes a fixed number of random samples from each class, we perform HBR on $\mathcal{D}^{\text{CLP}}$ with $\alpha=1$.

\textbf{Case Study 2.} 
\textit{Design.} In this case study, we compute resampling ratios using AUM as the proxy for $\hat{h}_c$. We set $\alpha \in \{1, 3, 5\}$ for CIFAR-10 and $\alpha \in \{1, 2\}$ for CIFAR-100. Larger $\alpha$ values are used for CIFAR-10 to compensate for its weaker hardness imbalance relative to CIFAR-100. We find that setting $\alpha=3$ for CIFAR-100 leads to significant drop in overall performance although the reduction to class bias is also more substantial. Hence, the optimal $\alpha$ is likely somewhere between $2$ and $3$ for CIFAR-100, although finding it is beyond the scope of this work.

\textit{Oversampling strategies.} We employ three main oversampling strategies: (1) random oversampling, (2) SMOTE, and (3) synthetic samples generated using EDM \citep{karras2022elucidating}. Since training a diffusion model is computationally prohibitive, we use a million of synthetic samples made publicly available by \citet{wang2023better}. This is also why we do not use diffusion models in Case Study 1, as retraining them for each \textit{Scenario A} is significantly beyond the scope of this work. Because diffusion and generative models are known to preferentially generate easy samples (see Section \ref{sec2.3}), we further define three EDM sub-strategies:
\begin{enumerate}
    \item \textbf{rEDM (random EDM):} We simply pick synthetic images from the target class at random, without considering how hard or easy they are. This serves as a baseline to see whether any hardness-based selection is beneficial at all.
    \item \textbf{aEDM (average EDM):} For each class, we first compute the average confidence of the real training images. We then select synthetic images whose hardness scores are closest to that class average. The idea is to match the typical difficulty level of the class, neither too easy nor too hard.
    \item \textbf{hEDM (hardest EDM):} We select the synthetic images with the highest hardness scores. These samples are most likely to lie near class boundaries, in low-density or high-complexity regions.
\end{enumerate}

For both aEDM and hEDM, we first pre‑select a candidate pool per class: the $2b_c$ synthetic images whose confidence is closest to the class average (for aEDM) or the $2b_c$ hardest images (for hEDM), where $b_c$ is the number of additional samples needed to balance that class. During each resampling iteration, we then draw randomly from this pre‑selected pool. This ensures a sufficiently large and hardness‑targeted supply while maintaining stochasticity across training runs.

Throughout these definitions, we estimate hardness using the confidence of the ten baseline models—the same models used to compute AUM on real data. For real data, we could also use AUM, but for consistency we use confidence for both real and synthetic samples. We deliberately avoid AUM for synthetic samples because AUM relies on the evolution of the margin over training epochs; we lack this learning dynamics information for synthetic data without retraining all ten models from scratch on the combined real‑plus‑synthetic dataset. Using confidence from the already‑trained baseline models provides a practical, consistent, and well‑defined proxy for hardness on both real and synthetic data.

\textbf{Experimental setup} \textit{Datasets.} All experiments are performed on CIFAR-10 and CIFAR-100 using their standard train/test splits. No additional validation set is used. Images are normalized using dataset-specific channel means and standard deviations. During training we apply standard augmentations: random horizontal flips and $32\times32$ random crops with $4$-pixel padding.

\textit{Model and optimization.} We use a ResNet-18 variant adapted for low-resolution inputs: the original $7\times7$ convolution and initial max-pooling are replaced with a single $3\times3$ convolution. All models are trained with stochastic gradient descent (SGD) with momentum $0.9$, initial learning rate $0.1$, and weight decay $5\times10^{-4}$. The learning rate is multiplied by $0.2$ at epochs $60$, $120$, and $160$. Training runs for $200$ epochs with batch size $128$.

\textit{Implementation notes.} All experiments are implemented in PyTorch.

\textit{Controlling data imbalance.} For the choice of $f$ in Equation \ref{equation 4.1}, we use the identity function, as it is intuitive that a class twice as hard should receive twice as many samples to address the imbalance. While this choice is guided by intuition, we acknowledge that other functional forms may be more appropriate, possibly in a dataset-dependent manner. A systematic investigation of this question is left for future work.

\textit{Evaluation.} For each trained model we compute class-level Recall and Precision on the standard test set. From these per-class scores we compute the fairness metrics defined in Section~\ref{sec3.2} (Eqs.\ for $\Delta_m$, $\Delta_q$, $\Delta_{he}$, $\Delta_\sigma$, and $\Delta_{\text{MAD}}$).

\section{Ensuring statistical reliability and reproducibility}\label{secA2}

Model performance in both case studies depends on stochastic components arising from data pruning and resampling. To ensure robustness and enable paired statistical testing, we adopt a deterministic seed scheme that guarantees reproducibility and one-to-one correspondence between models across datasets.

In Case Study~1, the source of randomness differs between the base and resampled datasets. For models trained on the base pruned datasets ($\mathcal{D}^{CLP}$), randomness originates solely from the pruning process. For models trained on the corresponding resampled datasets, randomness arises from both pruning and resampling. Hence, to ensure robust results we generate four independently pruned datasets for each pruning threshold, each using a distinct pruning seed, and train four independently initialized models on each of them (i.e., four pruned datasets × four models each). Model initialization seeds are computed deterministically as
\[
\text{seed}_{\text{model}}(i,j) = 42 + 420{,}000 \cdot i + 42 \cdot j,
\]
where $i \in \{0, 1, 2, 3\}$ indexes the dataset replicate and $j \in \{0, 1, 2, 3\}$ indexes the model within that replicate. These model seeds are independent of both the pruning and resampling seeds. Meanwhile, the pruning seed used to obtain $\mathcal{D}^{CLP}$ is set as
\[
\text{seed}_{\text{prune}}(i) = 42 \cdot i.
\]

In Case Study~2, randomness stems exclusively from resampling. Most oversampling and undersampling methods (except deterministic SMOTE) yield different datasets when initialized with different seeds. Accordingly, we again generate four independently resampled datasets, each controlled by its own resampling seed, and train four independent models per dataset using the same deterministic model-seed scheme as above. This design yields a total of $4 \times 4 = 16$ model-dataset combinations in both case studies.

\section{Design of paired t-test}\label{secA3}

To move beyond qualitative observations, and to enable consistent comparison across both case studies, we perform paired t-tests. Let $X_i$ denote the fairness metric from model $i$ trained on the base dataset ($\mathcal{D}^{\text{CLP}}$ in Case Study~1, and the full dataset in Case Study~2), and let $Y_i$ denote the same metric from model $i$ trained on the corresponding resampled dataset. We form $n=16$ pairs $(X_i, Y_i)$ and define $D_i = Y_i - X_i$.

The null hypothesis is $H_0: \mu_D = 0$, i.e., resampling has no effect. The one-sided alternative is $H_A: \mu_D > 0$, i.e., resampling improves fairness. We verify whether the null hypothesis can be rejected by applying Student's t-test, defined as follows:
\begin{align}
    t = \frac{\bar{D}}{s_D / \sqrt{n}},
    \label{equation 4.3}
\end{align}
with
\begin{align}
    s^2_D = \frac{1}{n - 1} \sum_{i-1}^n\left( D_i - \bar{D} \right)^2.
\end{align}
A $t$ value of $2$ indicates that observed average improvement to a specific fairness metric is twice as large as what we'd expect from sampling noise. In other words, the higher the $t$ value the larger and/or more consistent the improvements to fairness as per specific fairness metric. Hence, any negative values of $t$ indicate that applying HBR worsened fairness rather than improving it. After that, we compute the p-value, which gives the probability of observing a $t$ statistic at least as extreme as the one obtained, under the null hypothesis. In general, a p-value above $0.05$ indicates that null hypothesis cannot be rejected due to lack of statistical significance. 

\section{Results of paired t-test}\label{secA4}

\begin{table}[ht!]
    \centering
    \caption{\textbf{Case Study 1 (CIFAR-10):} t-statistics for gap- and dispersion-based metrics under different oversampling strategies. Significant results $(p < 0.05)$ are \textbf{bolded}. Positive (negative) values indicate that the observed reduction (increase) to class bias through HBR was statistically significant.}
    \begin{tabular}{lc|cc|cc|cc}
        \toprule
            \makecell{Fairness\\metric} & \makecell{Pruning\\rate} & \multicolumn{2}{c|}{random oversampling} & \multicolumn{2}{c|}{SMOTE oversampling} & \multicolumn{2}{c}{holdout oversampling} \\
             & & Recall & Precision & Recall & Precision & Recall & Precision \\
        \midrule
            \multirow{6}{*}{$\Delta_{m}$}
             & 33 & 0.17 & -0.52 & 1.82 & -1.57 & \textbf{4.52} & \textbf{-3.93} \\
             & 42 & -0.93 & -0.12 & -0.36 & -0.59 & \textbf{3.24} & -1.33 \\
             & 50 & -1.41 & 1.02 & 0.73 & 0.20 & \textbf{5.37} & -1.75 \\
             & 58 & -2.09 & 1.60 & -0.22 & -0.70 & 1.63 & -0.82 \\
             & 66 & -0.77 & 0.98 & -0.50 & -0.65 & \textbf{2.51} & \textbf{-2.89} \\
             & 75 & -0.61 & 0.40 & 0.12 & -0.26 & \textbf{3.24} & -0.76 \\
        \midrule
            \multirow{6}{*}{$\Delta_{q}$}
             & 33 & 0.89 & -1.38 & 1.77 & \textbf{-2.42} & \textbf{5.00} & \textbf{-3.81} \\
             & 42 & -0.91 & -0.76 & -0.11 & -1.45 & \textbf{5.49} & -1.11 \\
             & 50 & -0.10 & -0.71 & 1.70 & -0.37 & \textbf{6.20} & -2.09 \\
             & 58 & -1.92 & -0.57 & -0.69 & -1.01 & \textbf{2.89} & -0.69 \\
             & 66 & -0.74 & -0.45 & -0.16 & -1.40 & \textbf{2.29} & \textbf{-2.38} \\
             & 75 & -0.18 & -1.27 & 0.87 & \textbf{-3.54} & \textbf{5.08} & \textbf{-2.29} \\
        \midrule
            \multirow{6}{*}{$\Delta_{he}$}
             & 33 & \textbf{2.15} & -1.53 & \textbf{3.03} & \textbf{-2.61} & \textbf{5.16} & \textbf{-2.90} \\
             & 42 & -0.10 & 0.59 & 1.29 & -1.61 & \textbf{6.63} & -0.94 \\
             & 50 & 1.63 & -1.53 & \textbf{2.92} & -1.15 & \textbf{6.09} & -1.33 \\
             & 58 & -0.33 & -0.76 & 1.31 & -1.87 & \textbf{3.72} & -0.96 \\
             & 66 & 1.40 & 1.04 & \textbf{2.72} & -0.74 & \textbf{6.52} & -2.04 \\
             & 75 & 1.22 & -0.11 & \textbf{2.54} & \textbf{-2.16} & \textbf{9.22} & -1.58 \\
        \midrule
            \multirow{6}{*}{$\Delta_{\sigma}$}
             & 33 & 0.97 & -1.21 & 1.97 & \textbf{-2.27} & \textbf{4.82} & \textbf{-4.64} \\
             & 42 & -0.42 & -0.86 & 0.40 & -1.48 & \textbf{6.03} & -1.44 \\
             & 50 & -0.34 & -1.18 & 1.23 & -0.56 & \textbf{5.93} & -1.99 \\
             & 58 & -1.89 & -0.56 & -0.41 & -1.51 & \textbf{2.46} & -0.72 \\
             & 66 & -0.69 & -0.32 & 0.28 & -1.36 & \textbf{3.81} & \textbf{-2.68} \\
             & 75 & 0.14 & -1.60 & 0.91 & \textbf{-2.27} & \textbf{5.77} & -1.88 \\
        \midrule
            \multirow{6}{*}{$\Delta_{MAD}$}
             & 33 & 0.12 & -0.48 & 1.17 & -1.71 & 1.07 & 0.16 \\
             & 42 & -1.30 & -0.04 & -1.21 & -1.14 & -1.57 & 1.30 \\
             & 50 & -1.73 & \textbf{-3.85} & -0.26 & \textbf{-5.91} & -0.36 & \textbf{-2.17} \\
             & 58 & -0.74 & -1.92 & -2.08 & -0.76 & -0.00 & 0.35 \\
             & 66 & 0.34 & -1.43 & 0.41 & -1.44 & -0.30 & -0.70 \\
             & 75 & -1.45 & \textbf{-2.49} & -0.92 & -0.72 & -0.98 & 0.64 \\
        \bottomrule
    \end{tabular}
\end{table}

\begin{table}[ht!]
    \centering
    \caption{\textbf{Case Study 1 (CIFAR-100):} t-statistics for gap- and dispersion-based metrics under different oversampling strategies. Significant results $(p < 0.05)$ are \textbf{bolded}.}
    \begin{tabular}{lc|cc|cc|cc}
        \toprule
            \makecell{Fairness\\metric} & \makecell{Pruning\\rate} & \multicolumn{2}{c|}{random oversampling} & \multicolumn{2}{c|}{SMOTE oversampling} & \multicolumn{2}{c}{holdout oversampling} \\
             & & Recall & Precision & Recall & Precision & Recall & Precision \\
        \midrule
            \multirow{6}{*}{$\Delta_{m}$}
             & 33 & 0.99 & \textbf{-3.02} & 1.95 & \textbf{-4.61} & \textbf{4.73} & \textbf{-3.83} \\
             & 42 & \textbf{3.32} & -0.62 & \textbf{2.33} & -1.05 & \textbf{6.40} & \textbf{-2.29} \\
             & 50 & 1.48 & \textbf{-3.43} & 1.44 & \textbf{-3.33} & \textbf{8.68} & \textbf{-3.03} \\
             & 58 & -1.14 & -2.12 & -0.00 & -2.01 & \textbf{5.09} & \textbf{-4.07} \\
             & 66 & \textbf{6.15} & -1.54 & \textbf{3.54} & -1.74 & \textbf{11.41} & \textbf{-3.54} \\
             & 75 & 1.24 & \textbf{-2.86} & 1.50 & \textbf{-3.91} & \textbf{8.86} & \textbf{-5.47} \\
        \midrule
            \multirow{6}{*}{$\Delta_{q}$}
             & 33 & \textbf{2.54} & \textbf{-4.99} & \textbf{3.18} & \textbf{-4.80} & \textbf{12.17} & \textbf{-6.24} \\
             & 42 & \textbf{5.09} & \textbf{-4.09} & \textbf{4.86} & \textbf{-4.31} & \textbf{10.49} & \textbf{-7.44} \\
             & 50 & \textbf{4.19} & \textbf{-4.65} & \textbf{4.58} & \textbf{-4.32} & \textbf{10.98} & \textbf{-4.86} \\
             & 58 & \textbf{4.82} & \textbf{-3.57} & \textbf{7.45} & \textbf{-3.60} & \textbf{18.91} & \textbf{-4.57} \\
             & 66 & \textbf{12.54} & -1.97 & \textbf{11.23} & \textbf{-2.55} & \textbf{18.24} & \textbf{-3.19} \\
             & 75 & \textbf{7.72} & \textbf{-4.11} & \textbf{7.29} & \textbf{-6.41} & \textbf{21.04} & \textbf{-6.22} \\
        \midrule
            \multirow{6}{*}{$\Delta_{he}$}
             & 33 & \textbf{9.10} & \textbf{-7.61} & \textbf{8.98} & \textbf{-7.97} & \textbf{17.57} & \textbf{-9.58} \\
             & 42 & \textbf{9.61} & \textbf{-6.00} & \textbf{7.73} & \textbf{-9.19} & \textbf{13.90} & \textbf{-12.86} \\
             & 50 & \textbf{8.01} & \textbf{-5.01} & \textbf{12.06} & \textbf{-6.26} & \textbf{16.80} & \textbf{-8.43} \\
             & 58 & \textbf{11.04} & \textbf{-4.70} & \textbf{11.33} & \textbf{-2.83} & \textbf{23.33} & \textbf{-4.06} \\
             & 66 & \textbf{13.53} & \textbf{-4.89} & \textbf{11.09} & \textbf{-4.61} & \textbf{22.57} & \textbf{-8.69} \\
             & 75 & \textbf{20.60} & \textbf{-5.75} & \textbf{12.13} & \textbf{-8.14} & \textbf{26.02} & \textbf{-10.53} \\
        \midrule
            \multirow{6}{*}{$\Delta_{\sigma}$}
             & 33 & \textbf{3.41} & \textbf{-6.99} & \textbf{4.28} & \textbf{-6.23} & \textbf{15.18} & \textbf{-7.25} \\
             & 42 & \textbf{4.82} & \textbf{-6.14} & \textbf{4.87} & \textbf{-5.85} & \textbf{10.20} & \textbf{-10.66} \\
             & 50 & \textbf{7.06} & \textbf{-5.26} & \textbf{6.67} & \textbf{-5.56} & \textbf{15.11} & \textbf{-6.09} \\
             & 58 & \textbf{5.76} & \textbf{-3.94} & \textbf{9.60} & \textbf{-3.49} & \textbf{16.25} & \textbf{-5.09} \\
             & 66 & \textbf{11.13} & -1.24 & \textbf{12.27} & -1.63 & \textbf{22.35} & \textbf{-2.97} \\
             & 75 & \textbf{8.36} & \textbf{-5.50} & \textbf{6.37} & \textbf{-7.11} & \textbf{19.86} & \textbf{-7.82} \\
        \midrule
            \multirow{6}{*}{$\Delta_{MAD}$}
             & 33 & 2.08 & \textbf{-4.47} & 1.37 & \textbf{-3.69} & \textbf{4.12} & \textbf{-3.13} \\
             & 42 & \textbf{3.35} & \textbf{-3.20} & 1.29 & \textbf{-3.98} & \textbf{4.48} & \textbf{-3.04} \\
             & 50 & 0.15 & \textbf{-3.72} & 1.62 & \textbf{-2.37} & \textbf{4.81} & \textbf{-2.82} \\
             & 58 & 1.85 & -1.83 & \textbf{4.26} & -1.77 & \textbf{4.65} & \textbf{-3.04} \\
             & 66 & \textbf{4.53} & -0.21 & \textbf{6.16} & -0.13 & \textbf{7.93} & -0.16 \\
             & 75 & \textbf{7.91} & \textbf{-2.64} & \textbf{2.95} & \textbf{-2.93} & \textbf{5.23} & -1.91 \\
        \bottomrule
    \end{tabular}
\end{table}

\textbf{Changes to class bias are not restricted to the most extreme classes.} We find that the relative magnitudes of changes to class bias do not directly align with their statistical significance. While HBR yields the largest absolute reductions in $\Delta_m$, followed by $\Delta_q$ and $\Delta_{he}$, the corresponding $t$-values show the reverse trend. This discrepancy suggests that improvements in the recall gap across most extreme classes ($\Delta_m$ and $\Delta_q$) are more variable across runs, but also more pronounced. Meanwhile, improvements spread across all hard and easy classes ($\Delta_{he}$) are smaller but more consistent. This can be explained by higher robustness of $\Delta_{he}$ as a fairness metric, as we discussed in Section \ref{sec3.2}. Paired with the fact that $\Delta_{MAD}$ gets reduced on CIFAR-100, this indicates that \textit{the impact of HBR is not confined to the most extreme classes but rather spread across the whole hardness spectrum}. We believe that $\Delta_{MAD}$ increasing on CIFAR-10 as a result of HBR does not contradict these insights but rather is a consequence of the unique hardness spectrum characterizing this dataset. As is visible in Figure \ref{fig:fig1}, while on CIFAR-100 recall decreases almost linearly across classes, CIFAR-10 recall follows a logistic-like trend, with only a few low-performing classes and most classes clustered around similar recall values. This means that on CIFAR-10 the effects of HBR are more related to the hardest classes, and any changes to the $\Delta_{MAD}$ are bound to be insignificant and inconsistent—if most classes already have similar recall, there is little room for median dispersion to change.

\setlength{\tabcolsep}{2pt}      %
\renewcommand{\arraystretch}{1.1}

\begin{sidewaystable}[p]
    \centering
    \caption{\textbf{Case Study 2 (CIFAR-10):} t-statistics for gap- and dispersion-based metrics under different oversampling strategies. Significant results $(p < 0.05)$ are \textbf{bolded}.}
    \begin{tabular}{l l|ccc|ccc|ccc|ccc|ccc}
        \toprule
            Metric & Oversampling & \multicolumn{3}{c|}{$\Delta_{m}$} & \multicolumn{3}{c|}{$\Delta_{q}$} & \multicolumn{3}{c|}{$\Delta_{he}$} & \multicolumn{3}{c|}{$\Delta_{\sigma}$} & \multicolumn{3}{c}{$\Delta_{MAD}$} \\
            &  & $\alpha=1$ & $\alpha=3$ & $\alpha=5$ & $\alpha=1$ & $\alpha=3$ & $\alpha=5$ & $\alpha=1$ & $\alpha=3$ & $\alpha=5$ & $\alpha=1$ & $\alpha=3$ & $\alpha=5$ & $\alpha=1$ & $\alpha=3$ & $\alpha=5$ \\
        \midrule
            \multirow{5}{*}{Recall} & random & -0.38 & 1.83 & 1.90 & -0.56 & 1.82 & \textbf{2.40} & -0.15 & 1.83 & \textbf{2.82} & -0.39 & 2.03 & \textbf{2.86} & -0.78 & -0.27 & -0.56 \\
            & SMOTE & 0.21 & 0.29 & -0.73 & 0.09 & 0.94 & 0.69 & 1.61 & 1.15 & 1.81 & 0.89 & 1.39 & 0.90 & 0.20 & \textbf{-2.26} & 0.75 \\
            & rEDM & 1.96 & \textbf{4.11} & \textbf{5.69} & 1.58 & \textbf{4.39} & \textbf{4.69} & 1.96 & \textbf{3.91} & \textbf{5.05} & \textbf{2.84} & \textbf{5.59} & \textbf{6.67} & -1.89 & 0.54 & 0.24 \\
            & aEDM & \textbf{2.21} & \textbf{5.90} & \textbf{7.29} & 1.88 & \textbf{5.76} & \textbf{6.14} & \textbf{2.35} & \textbf{5.16} & \textbf{7.66} & \textbf{3.12} & \textbf{6.89} & \textbf{7.18} & -1.63 & -0.51 & 1.22 \\
            & hEDM & \textbf{3.22} & \textbf{7.04} & \textbf{10.77} & \textbf{6.34} & \textbf{9.90} & \textbf{10.35} & \textbf{5.75} & \textbf{17.71} & \textbf{12.31} & \textbf{6.05} & \textbf{13.61} & \textbf{14.83} & -0.12 & -0.18 & 0.08 \\
        \midrule
            \multirow{5}{*}{Precision} & random & 1.50 & -0.37 & 0.36 & 1.85 & 0.26 & 1.12 & -0.65 & -1.38 & -0.98 & 1.46 & -0.09 & 0.68 & -2.11 & -1.67 & -1.74 \\
            & SMOTE & -1.97 & -1.41 & -1.57 & -1.45 & -0.97 & -1.80 & \textbf{-2.93} & \textbf{-3.03} & \textbf{-5.61} & -1.55 & -1.51 & \textbf{-2.31} & \textbf{-2.14} & -0.69 & -1.59 \\
            & rEDM & -1.33 & -1.84 & \textbf{-2.65} & -0.48 & -0.67 & 0.67 & -1.96 & -1.63 & \textbf{-2.47} & -1.05 & -0.98 & -0.86 & -1.33 & 0.01 & -0.51 \\
            & aEDM & \textbf{-2.68} & -1.47 & \textbf{-2.94} & -1.39 & -0.15 & -0.64 & \textbf{-2.75} & -1.63 & -1.36 & \textbf{-2.14} & -0.69 & -1.49 & -0.50 & -0.19 & 0.24 \\
            & hEDM & -1.79 & \textbf{-3.51} & \textbf{-3.37} & -1.50 & \textbf{-2.46} & -1.95 & \textbf{-3.21} & \textbf{-4.58} & \textbf{-2.55} & -1.65 & \textbf{-3.25} & -1.93 & -0.32 & -1.25 & -0.98 \\
        \bottomrule
    \end{tabular}
    \bigskip \bigskip
    \caption{\textbf{Case Study 2 (CIFAR-100):} t-statistics for gap- and dispersion-based metrics under different oversampling strategies. Significant results $(p < 0.05)$ are \textbf{bolded}.}
    \begin{tabular}{l l|cc|cc|cc|cc|cc}
        \toprule
            Metric & Oversampling & \multicolumn{2}{c|}{$\Delta_{m}$} & \multicolumn{2}{c|}{$\Delta_{q}$} & \multicolumn{2}{c|}{$\Delta_{he}$} & \multicolumn{2}{c|}{$\Delta_{\sigma}$} & \multicolumn{2}{c}{$\Delta_{MAD}$} \\
            &  & $\alpha=1$ & $\alpha=2$ & $\alpha=1$ & $\alpha=2$ & $\alpha=1$ & $\alpha=2$ & $\alpha=1$ & $\alpha=2$ & $\alpha=1$ & $\alpha=2$ \\
        \midrule
            \multirow{5}{*}{Recall} & random & 0.68 & 1.91 & -0.56 & 0.83 & 1.93 & \textbf{2.81} & 0.47 & \textbf{2.27} & 0.22 & 1.50 \\
            & SMOTE & -0.76 & 0.23 & -1.01 & 0.46 & 0.55 & \textbf{4.29} & -1.05 & 1.45 & -0.75 & 0.20 \\
            & rEDM & 0.49 & 0.17 & \textbf{2.95} & \textbf{8.10} & \textbf{5.48} & \textbf{11.61} & \textbf{3.77} & \textbf{13.19} & 0.89 & \textbf{3.58} \\
            & aEDM & 0.55 & 0.33 & 1.58 & \textbf{4.68} & \textbf{2.14} & \textbf{9.08} & \textbf{2.49} & \textbf{7.21} & 0.72 & 1.58 \\
            & hEDM & \textbf{2.35} & \textbf{2.54} & \textbf{3.99} & \textbf{9.82} & \textbf{9.94} & \textbf{14.03} & \textbf{7.31} & \textbf{12.24} & \textbf{3.30} & \textbf{6.26} \\
        \midrule
            \multirow{5}{*}{Precision} & random & 0.08 & -1.75 & \textbf{-2.74} & \textbf{-3.42} & \textbf{-3.61} & \textbf{-4.30} & \textbf{-2.90} & \textbf{-4.42} & 0.35 & \textbf{-2.79} \\
            & SMOTE & -1.89 & \textbf{-3.49} & \textbf{-3.57} & \textbf{-7.68} & \textbf{-3.88} & \textbf{-6.61} & \textbf{-3.58} & \textbf{-7.72} & -0.28 & \textbf{-2.49} \\
            & rEDM & -1.24 & \textbf{-2.48} & -2.01 & \textbf{-4.26} & \textbf{-3.73} & \textbf{-6.09} & \textbf{-2.43} & \textbf{-5.84} & -0.01 & -0.47 \\
            & aEDM & -0.81 & -1.32 & -1.82 & \textbf{-3.74} & \textbf{-4.18} & \textbf{-4.64} & \textbf{-2.65} & \textbf{-4.56} & 0.18 & -1.38 \\
            & hEDM & \textbf{-3.35} & \textbf{-4.98} & \textbf{-5.61} & \textbf{-8.55} & \textbf{-7.15} & \textbf{-8.97} & \textbf{-6.86} & \textbf{-10.16} & -0.51 & \textbf{-2.62} \\
        \bottomrule
    \end{tabular}
\end{sidewaystable}

\setlength{\tabcolsep}{6pt}      %
\renewcommand{\arraystretch}{1}

\begin{table}[ht!]
    \centering
    \caption{\textbf{Case Study 1:} t-statistics for $\Delta_{avg}$ under different oversampling strategies. Significant results $(p<0.05)$ are \textbf{bolded}.}
    \begin{tabular}{lc|ccc|ccc}
        \toprule
            \makecell{Metric} & \makecell{Pruning\\rate} & \multicolumn{3}{c|}{CIFAR-10} & \multicolumn{3}{c}{CIFAR-100} \\
             & & random & SMOTE & holdout & random & SMOTE & holdout\\
        \midrule
        \multirow{6}{*}{Recall}
             & 33 & \textbf{-2.30} & -0.54 & 0.07 & \textbf{-12.87} & \textbf{-12.14} & \textbf{-3.68} \\
             & 42 & -1.12 & -1.95 & 0.17 & \textbf{-14.20} & \textbf{-12.68} & \textbf{-3.42} \\
             & 50 & -1.54 & -0.91 & 1.43 & \textbf{-9.53} & \textbf{-9.51} & \textbf{-3.57} \\
             & 58 & -1.82 & -1.55 & 0.83 & \textbf{-14.23} & \textbf{-12.74} & \textbf{-4.50}\\
             & 66 & -1.48 & -1.57 & 0.25 & \textbf{5.52} & \textbf{6.17} & \textbf{14.51} \\
             & 75 & -1.41 & -1.46 & 0.95 & \textbf{-8.28} & \textbf{-10.74} & \textbf{-4.46} \\
        \midrule
            \multirow{6}{*}{Precision}
             & 33 & \textbf{-2.17} & -0.26 & 0.53 & \textbf{-9.65} & \textbf{-9.44} & 0.72 \\
             & 42 & -1.08 & -1.88 & 0.40 & \textbf{-9.26} & \textbf{-11.05} & \textbf{2.53} \\
             & 50 & -1.49 & -0.80 & 2.03 & \textbf{-8.02} & \textbf{-7.83} & \textbf{3.13} \\
             & 58 & -1.88 & -1.52 & 0.95 & \textbf{-9.89} & \textbf{-8.36} & 0.67 \\
             & 66 & -1.44 & -1.43 & 0.73 & \textbf{7.76} & \textbf{8.05} & \textbf{19.42} \\
             & 75 & -1.34 & -1.32 & 1.38 & \textbf{-6.27} & \textbf{-7.86} & \textbf{4.02} \\
        \bottomrule
    \end{tabular}
    \bigskip\bigskip
    \centering
    \caption{\textbf{Case Study 2:} t-statistics for $\Delta_{avg}$ under different oversampling strategies. Significant results $(p<0.05)$ are \textbf{bolded}.}
    \begin{tabular}{l l|ccc|cc}
        \toprule
            Metric & Oversampling & \multicolumn{3}{c|}{CIFAR-10} & \multicolumn{2}{c}{CIFAR-100} \\
            & & $\alpha=1$ & $\alpha=3$ & $\alpha=5$ & $\alpha=1$ & $\alpha=2$ \\
        \midrule
            \multirow{5}{*}{Recall} & random & 1.25 & 0.37 & 0.10 & \textbf{-2.70} & \textbf{-7.25} \\
            & SMOTE & 0.45 & -1.66 & \textbf{-2.29} & \textbf{-4.50} & \textbf{-7.70} \\
            & rEDM & 0.94 & 1.05 & \textbf{2.96} & 0.01 & -1.06 \\
            & aEDM & 2.05 & \textbf{3.31} & \textbf{4.62} & 1.37 & -0.72 \\
            & hEDM & 1.41 & \textbf{6.96} & \textbf{7.04} & \textbf{2.67} & \textbf{3.30} \\
        \midrule
            \multirow{5}{*}{Precision} & random & 1.19 & 0.42 & 0.16 & \textbf{-2.32} & \textbf{-6.67} \\
            & SMOTE & 0.54 & -1.56 & \textbf{-2.18} & \textbf{-4.11} & \textbf{-6.05} \\
            & rEDM & 1.06 & 1.22 & \textbf{3.22} & 1.67 & 2.07 \\
            & aEDM & \textbf{2.18} & \textbf{3.51} & \textbf{4.98} & \textbf{2.31} & 1.52 \\
            & hEDM & 1.62 & \textbf{7.74} & \textbf{7.61} & \textbf{5.95} & \textbf{9.49} \\
        \bottomrule
    \end{tabular}
\end{table}

\clearpage

\section{Additional experimental results}\label{secA5}

In this section we include further experiments conducted for Section \ref{sec4}. Specifically, we show the changes to true positives, true negatives, false positives, and false negatives across classes as an effect of HBR for top-5 most extreme classes. These classes were chosen based on the recall averaged over the sixteen models trained on $\mathcal{D}^{CLP}$ with pruning rate of $50\%$. This means that only the five classes with the highest and lowest averaged recalls are displayed, with the former representing the easiest classes and the latter the hardest ones. We decided to use Recall instead of AUM to sort classes, as we consider it as ground truth for hardness (just like accuracy is often used as a ground truth for hardness in relevant literature).

\begin{figure}[h!]
    \includegraphics[width=\textwidth]{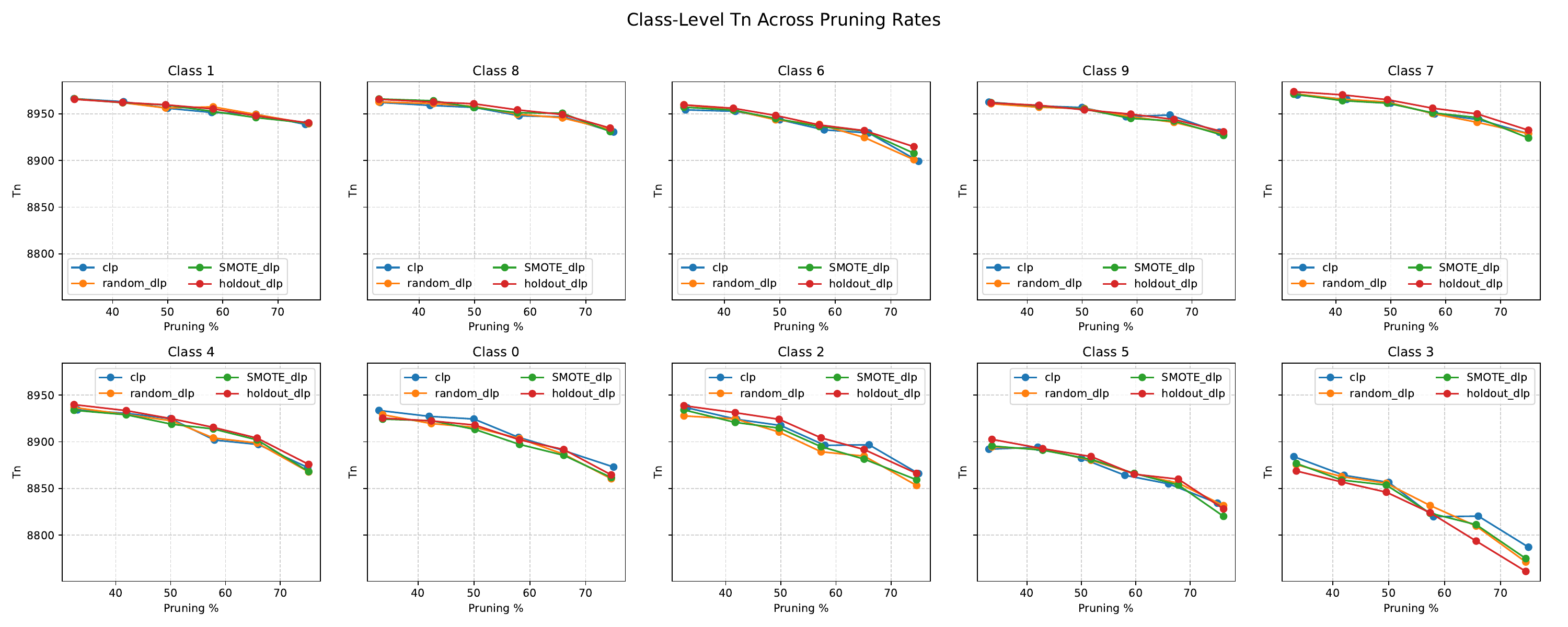}
    \caption{True negatives across classes on CIFAR-10.}
    \label{fig: CIFAR-10 Tn}
\end{figure}

\begin{figure}[h!]
    \includegraphics[width=\textwidth]{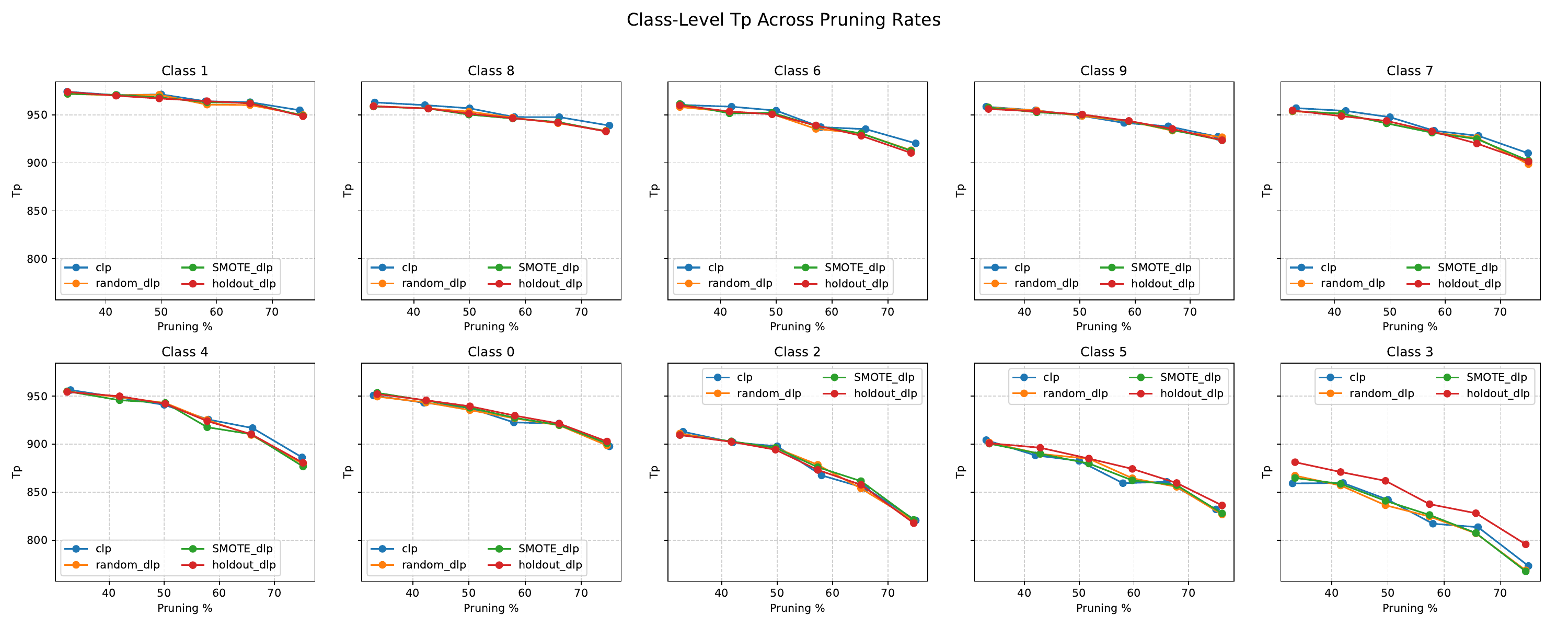}
    \caption{True positives across classes on CIFAR-10.}
    \label{fig: CIFAR-10 Tp}
\end{figure}

\begin{figure}[t!]
    \includegraphics[width=\textwidth]{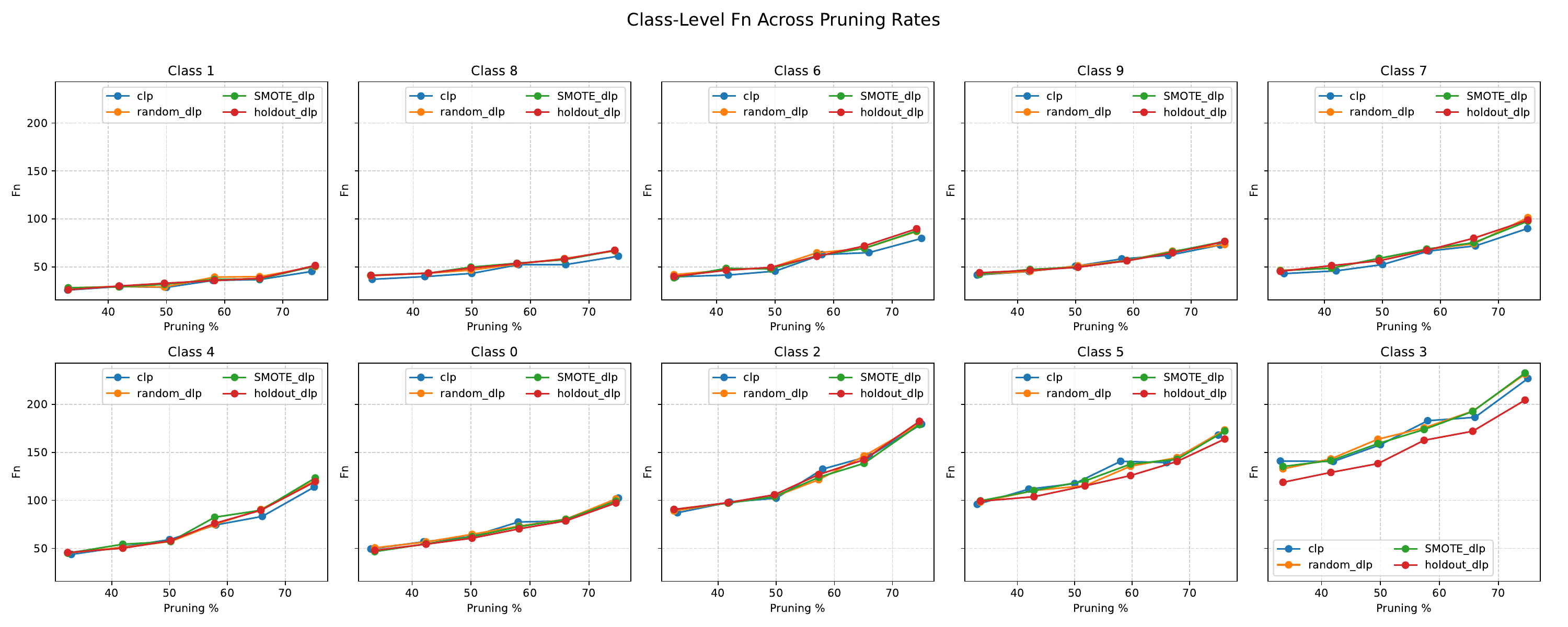}
    \caption{False negatives across classes on CIFAR-10.}
    \label{fig: CIFAR-10 Fn}
\end{figure}

\begin{figure}[t!]
    \includegraphics[width=\textwidth]{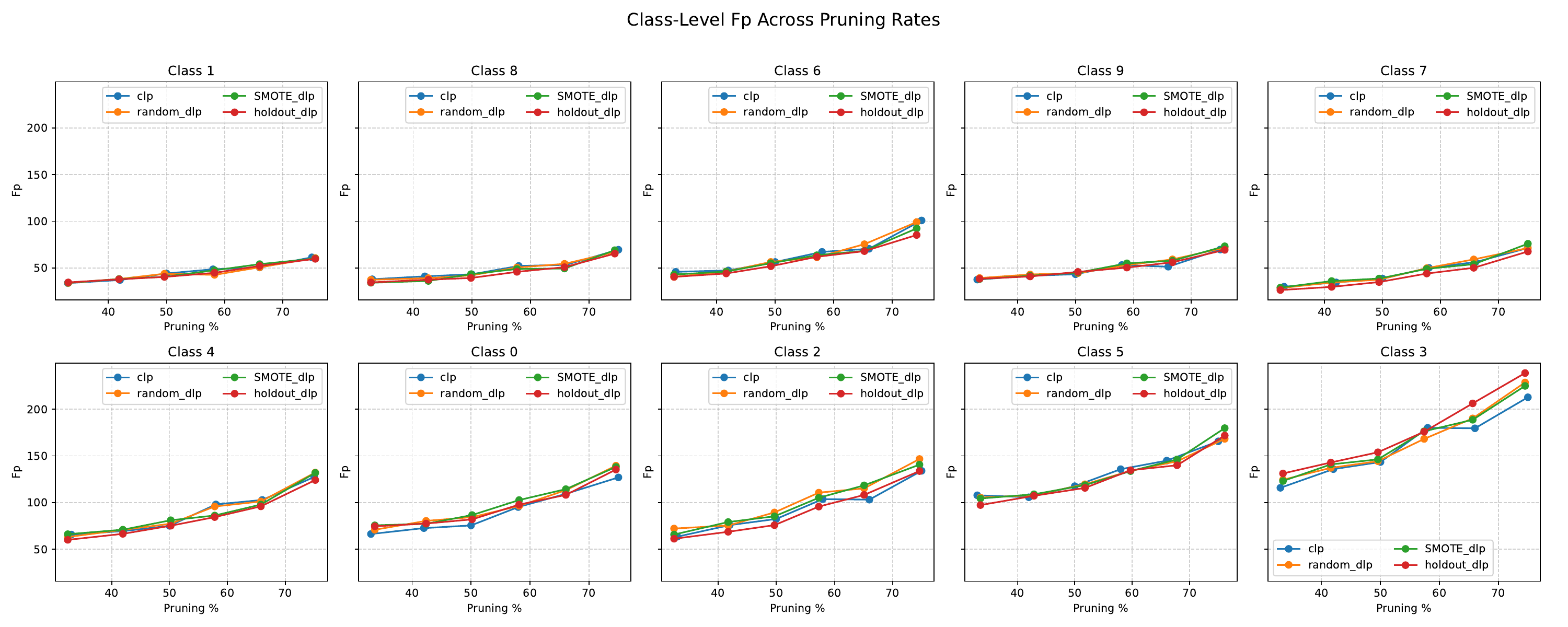}
    \caption{False positives across classes on CIFAR-10.}
    \label{fig: CIFAR-10 Fp}
\end{figure}

\begin{figure}[t!]
    \includegraphics[width=\textwidth]{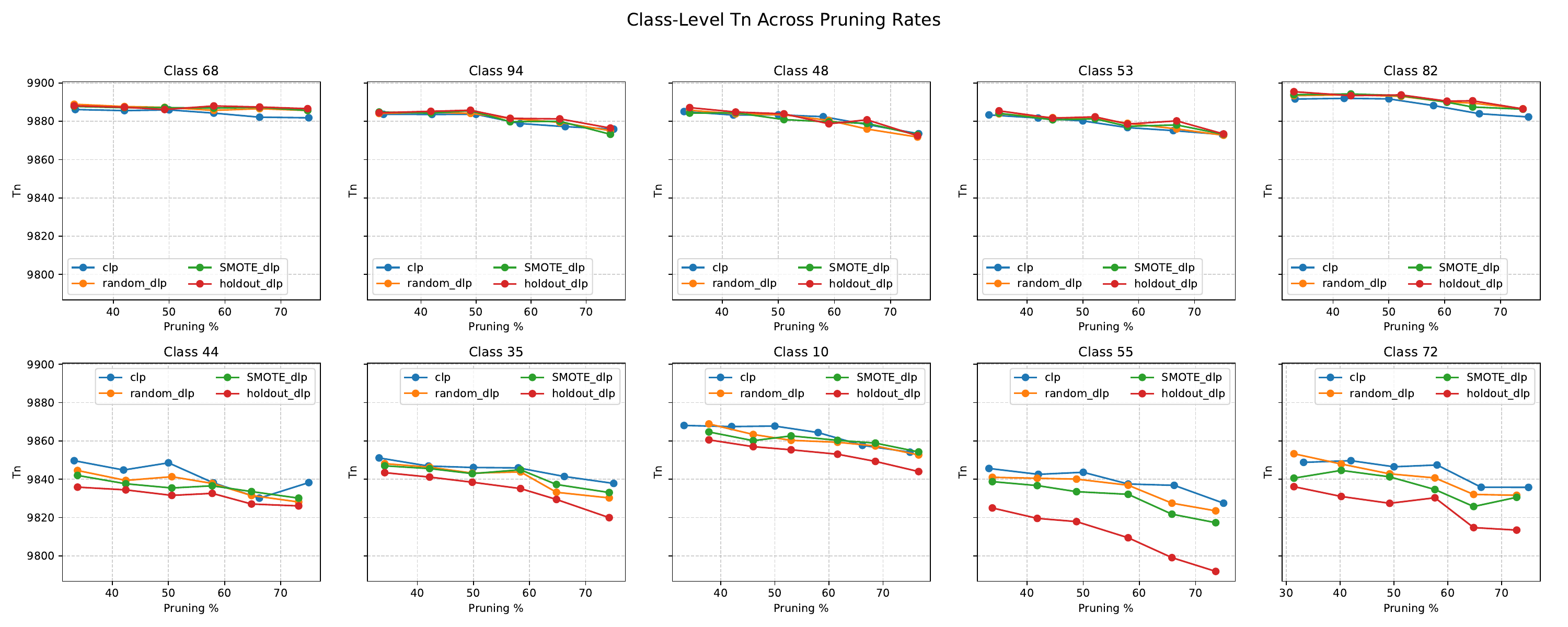}
    \caption{True negatives across classes on CIFAR-100.}
    \label{fig: CIFAR-100 Tn}
\end{figure}

\begin{figure}[t!]
    \includegraphics[width=\textwidth]{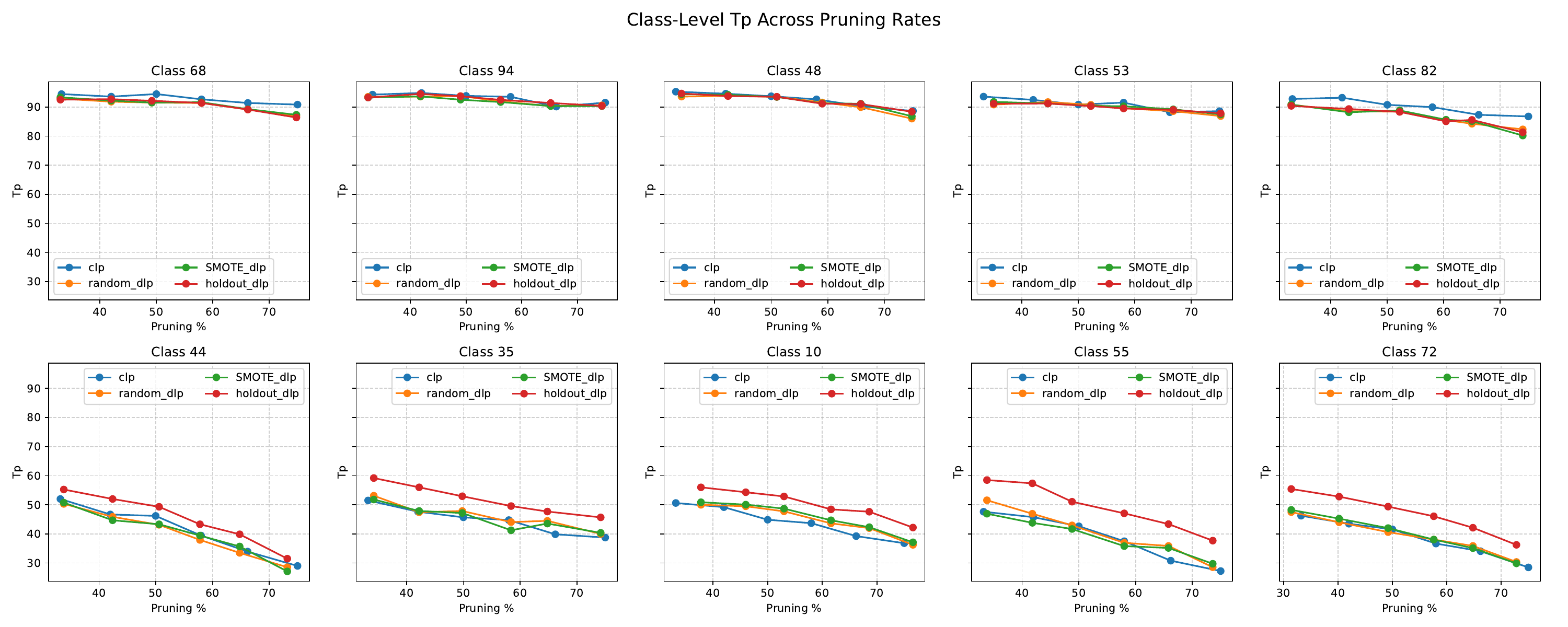}
    \caption{True positives across classes on CIFAR-100.}
    \label{fig: CIFAR-100 Tp}
\end{figure}

\begin{figure}[t!]
    \includegraphics[width=\textwidth]{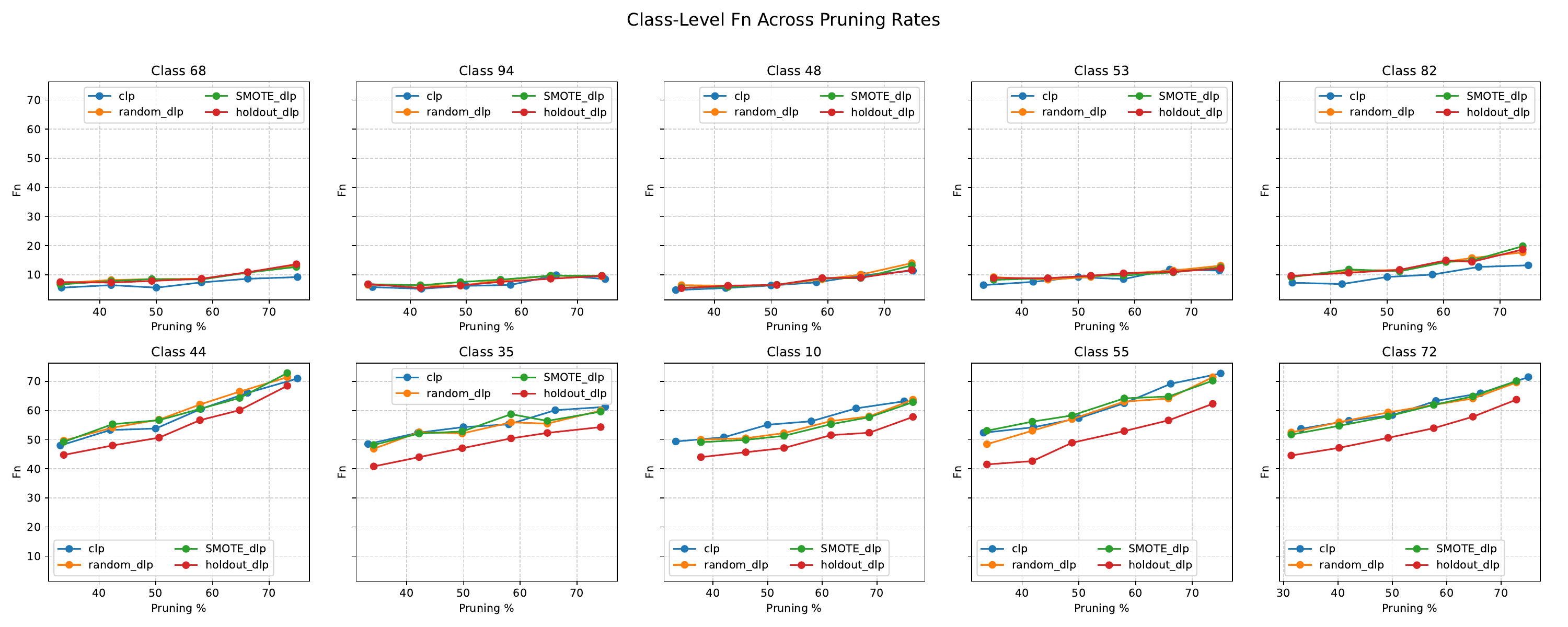}
    \caption{False negatives across classes on CIFAR-100.}
    \label{fig: CIFAR-100 Fn}
\end{figure}

\begin{figure}[t!]
    \includegraphics[width=\textwidth]{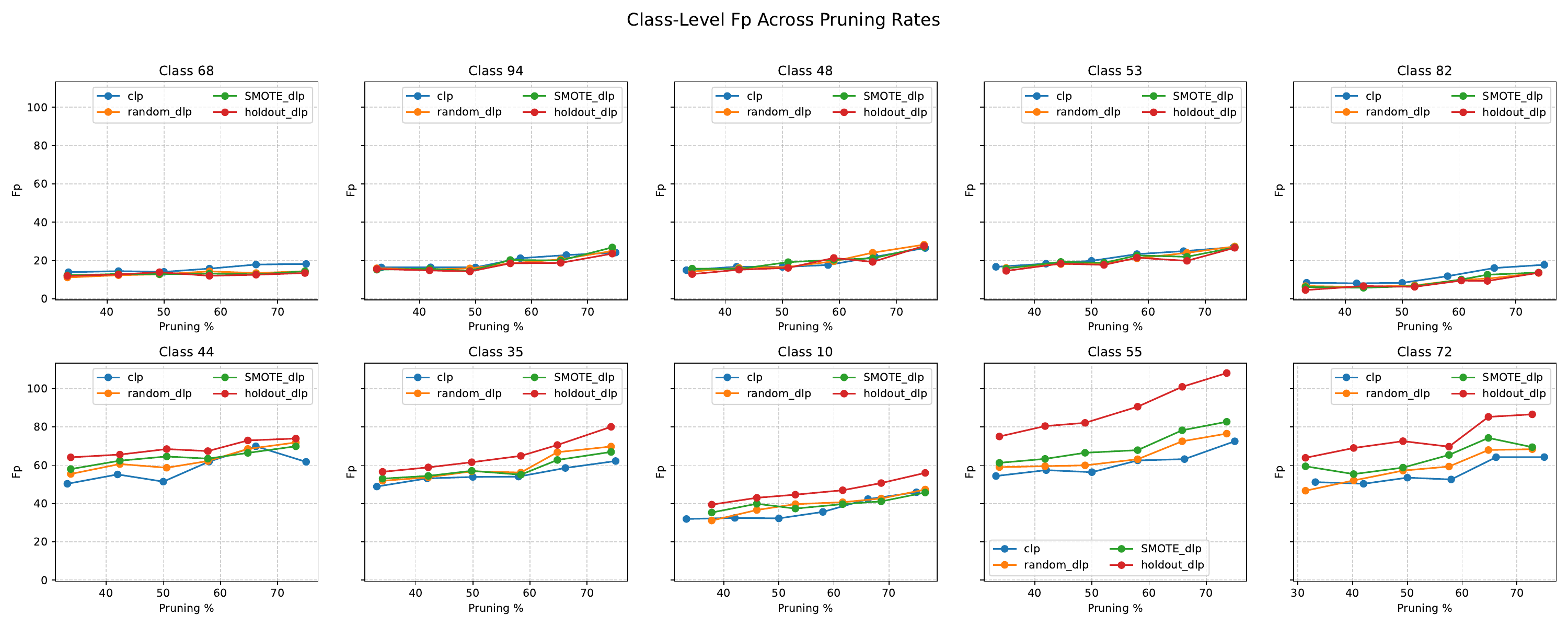}
    \caption{False positives across classes on CIFAR-100.}
    \label{fig: CIFAR-100 Fp}
\end{figure}

\clearpage

\end{appendices}

\section*{Declarations}

\noindent \textbf{Author Contributions} All authors contributed to the study conception and design. Material preparation, data collection and analysis were performed by Pawel Pukowski. The first draft of the manuscript was written by Pawel Pukowski and all authors commented on previous versions of the manuscript. All authors read and approved the final manuscript.

\noindent \textbf{Funding} This work was supported by the UK Research and Innovation (UKRI) Engineering and Physical Sciences Research Council (EPSRC) under the PhD Scholarship Grant.

\noindent \textbf{Data Availability} All the datasets are publicly available. Data descriptions are provided in Section \ref{sec: data_description}. The 1 million synthetic data generated by EDM was downloaded from \href{https://github.com/wzekai99/DM-Improves-AT}{https://github.com/wzekai99/DM-Improves-AT}

\noindent \textbf{Code Availability} The code will be made available after successful revision process.

\noindent \textbf{Conflict of interest} None.

\noindent \textbf{Ethical Approval} Not applicable

\noindent \textbf{Consent for Publication} All authors consent to the submission of this manuscript to the \textit{Machine Learning}

\bibliography{bibliography}

@inproceedings{zhu2026reducing,
    title={Reducing Class-Wise Performance Disparity via Margin Regularization},
    author={Beier Zhu and Kesen Zhao and Jiequan Cui and Qianru Sun and Yuan Zhou and Xun Yang and Hanwang Zhang},
    booktitle={The Fourteenth International Conference on Learning Representations},
    year={2026},
    url={https://openreview.net/forum?id=KfjpyOcPQj}
}

@article{brzezinski2024properties,
  title={Properties of fairness measures in the context of varying class imbalance and protected group ratios},
  author={Brzezinski, Dariusz and Stachowiak, Julia and Stefanowski, Jerzy and Szczech, Izabela and Susmaga, Robert and Aksenyuk, Sofya and Ivashka, Uladzimir and Yasinskyi, Oleksandr},
  journal={ACM Transactions on Knowledge Discovery from Data},
  volume={18},
  number={7},
  pages={1--18},
  year={2024},
  publisher={ACM New York, NY}
}

@article{li2025bias,
  title={Bias behind the wheel: Fairness testing of autonomous driving systems},
  author={Li, Xinyue and Chen, Zhenpeng and Zhang, Jie M and Sarro, Federica and Zhang, Ying and Liu, Xuanzhe},
  journal={ACM Transactions on Software Engineering and Methodology},
  volume={34},
  number={3},
  pages={1--24},
  year={2025},
  publisher={ACM New York, NY}
}

@article{katare2025analyzing,
  title={Analyzing and Mitigating Bias for Vulnerable Road Users by Addressing Class Imbalance in Datasets},
  author={Katare, Dewant and Noguero, David Solans and Park, Souneil and Kourtellis, Nicolas and Janssen, Marijn and Ding, Aaron Yi},
  journal={IEEE Open Journal of Intelligent Transportation Systems},
  year={2025},
  publisher={IEEE}
}

@article{rahman2013addressing,
  title={Addressing the class imbalance problem in medical datasets},
  author={Rahman, M Mostafizur and Davis, Darryl N},
  journal={International Journal of Machine Learning and Computing},
  volume={3},
  number={2},
  pages={224},
  year={2013},
  publisher={IACSIT Press}
}

@article{khushi2021comparative,
  title={A comparative performance analysis of data resampling methods on imbalance medical data},
  author={Khushi, Matloob and Shaukat, Kamran and Alam, Talha Mahboob and Hameed, Ibrahim A and Uddin, Shahadat and Luo, Suhuai and Yang, Xiaoyan and Reyes, Maranatha Consuelo},
  journal={IEEE Access},
  volume={9},
  pages={109960--109975},
  year={2021},
  publisher={IEEE}
}

@inproceedings{zong2022medfair,
  author={Yongshuo Zong and Yongxin Yang and Timothy M. Hospedales},
  title={MEDFAIR: Benchmarking Fairness for Medical Imaging},
  year={2023},
  cdate={1672531200000},
  url={https://openreview.net/forum?id=6ve2CkeQe5S},
  booktitle={ICLR}
}

@article{tasci2022bias,
  title={Bias and class imbalance in oncologic data—towards inclusive and transferrable AI in large scale oncology data sets},
  author={Tasci, Erdal and Zhuge, Ying and Camphausen, Kevin and Krauze, Andra V},
  journal={Cancers},
  volume={14},
  number={12},
  pages={2897},
  year={2022},
  publisher={MDPI}
}

@inproceedings{hashimoto2018fairness,
  title={Fairness without demographics in repeated loss minimization},
  author={Hashimoto, Tatsunori and Srivastava, Megha and Namkoong, Hongseok and Liang, Percy},
  booktitle={International Conference on Machine Learning},
  pages={1929--1938},
  year={2018},
  organization={PMLR}
}

@inproceedings{lin2017focal,
  title={Focal loss for dense object detection},
  author={Lin, Tsung-Yi and Goyal, Priya and Girshick, Ross and He, Kaiming and Doll{\'a}r, Piotr},
  booktitle={Proceedings of the IEEE international conference on computer vision},
  pages={2980--2988},
  year={2017}
}

@article{chawla2002smote,
  title={SMOTE: synthetic minority over-sampling technique},
  author={Chawla, Nitesh V and Bowyer, Kevin W and Hall, Lawrence O and Kegelmeyer, W Philip},
  journal={Journal of artificial intelligence research},
  volume={16},
  pages={321--357},
  year={2002}
}

@inproceedings{sinha2020class,
  title={Class-wise difficulty-balanced loss for solving class-imbalance},
  author={Sinha, Saptarshi and Ohashi, Hiroki and Nakamura, Katsuyuki},
  booktitle={Proceedings of the Asian conference on computer vision},
  year={2020}
}

@article{sinha2022class,
  title={Class-difficulty based methods for long-tailed visual recognition},
  author={Sinha, Saptarshi and Ohashi, Hiroki and Nakamura, Katsuyuki},
  journal={International Journal of Computer Vision},
  volume={130},
  number={10},
  pages={2517--2531},
  year={2022},
  publisher={Springer}
}

@inproceedings{he2016deep,
  title={Deep residual learning for image recognition},
  author={He, Kaiming and Zhang, Xiangyu and Ren, Shaoqing and Sun, Jian},
  booktitle={Proceedings of the IEEE conference on computer vision and pattern recognition},
  pages={770--778},
  year={2016}
}

@article{krizhevsky2009learning,
  title={Learning multiple layers of features from tiny images},
  author={Krizhevsky, Alex and Hinton, Geoffrey and others},
  year={2009},
  publisher={Toronto, ON, Canada}
}

@article{wang2021survey,
  title={A survey on curriculum learning},
  author={Wang, Xin and Chen, Yudong and Zhu, Wenwu},
  journal={IEEE transactions on pattern analysis and machine intelligence},
  volume={44},
  number={9},
  pages={4555--4576},
  year={2021},
  publisher={IEEE}
}

@article{soviany2022curriculum,
  title={Curriculum learning: A survey},
  author={Soviany, Petru and Ionescu, Radu Tudor and Rota, Paolo and Sebe, Nicu},
  journal={International Journal of Computer Vision},
  volume={130},
  number={6},
  pages={1526--1565},
  year={2022},
  publisher={Springer}
}

@article{settles2009active,
  title={Active learning literature survey},
  author={Settles, Burr},
  year={2009},
  publisher={University of Wisconsin-Madison Department of Computer Sciences}
}

@article{ren2021survey,
  title={A survey of deep active learning},
  author={Ren, Pengzhen and Xiao, Yun and Chang, Xiaojun and Huang, Po-Yao and Li, Zhihui and Gupta, Brij B and Chen, Xiaojiang and Wang, Xin},
  journal={ACM computing surveys (CSUR)},
  volume={54},
  number={9},
  pages={1--40},
  year={2021},
  publisher={ACM New York, NY}
}

@article{toneva2018empirical,
  title={An empirical study of example forgetting during deep neural network learning},
  author={Toneva, Mariya and Sordoni, Alessandro and Combes, Remi Tachet des and Trischler, Adam and Bengio, Yoshua and Gordon, Geoffrey J},
  journal={arXiv preprint arXiv:1812.05159},
  year={2018}
}

@article{paul2021deep,
  title={Deep learning on a data diet: Finding important examples early in training},
  author={Paul, Mansheej and Ganguli, Surya and Dziugaite, Gintare Karolina},
  journal={Advances in neural information processing systems},
  volume={34},
  pages={20596--20607},
  year={2021}
}

@article{sorscher2022beyond,
  title={Beyond neural scaling laws: beating power law scaling via data pruning},
  author={Sorscher, Ben and Geirhos, Robert and Shekhar, Shashank and Ganguli, Surya and Morcos, Ari},
  journal={Advances in Neural Information Processing Systems},
  volume={35},
  pages={19523--19536},
  year={2022}
}

@inproceedings{agarwal2022estimating,
  title={Estimating example difficulty using variance of gradients},
  author={Agarwal, Chirag and D'souza, Daniel and Hooker, Sara},
  booktitle={Proceedings of the IEEE/CVF Conference on Computer Vision and Pattern Recognition},
  pages={10368--10378},
  year={2022}
}

@article{yang2024generalized,
  title={Generalized out-of-distribution detection: A survey},
  author={Yang, Jingkang and Zhou, Kaiyang and Li, Yixuan and Liu, Ziwei},
  journal={International Journal of Computer Vision},
  volume={132},
  number={12},
  pages={5635--5662},
  year={2024},
  publisher={Springer}
}

@article{pleiss2020identifying,
  title={Identifying mislabeled data using the area under the margin ranking},
  author={Pleiss, Geoff and Zhang, Tianyi and Elenberg, Ethan and Weinberger, Kilian Q},
  journal={Advances in Neural Information Processing Systems},
  volume={33},
  pages={17044--17056},
  year={2020}
}

@article{maini2022characterizing,
  title={Characterizing datapoints via second-split forgetting},
  author={Maini, Pratyush and Garg, Saurabh and Lipton, Zachary and Kolter, J Zico},
  journal={Advances in Neural Information Processing Systems},
  volume={35},
  pages={30044--30057},
  year={2022}
}

@inproceedings{jia2023learning,
  title={Learning from training dynamics: Identifying mislabeled data beyond manually designed features},
  author={Jia, Qingrui and Li, Xuhong and Yu, Lei and Bian, Jiang and Zhao, Penghao and Li, Shupeng and Xiong, Haoyi and Dou, Dejing},
  booktitle={Proceedings of the AAAI Conference on Artificial Intelligence},
  volume={37},
  number={7},
  pages={8041--8049},
  year={2023}
}

@article{zhu2024exploring,
  title={Exploring the learning difficulty of data: Theory and measure},
  author={Zhu, Weiyao and Wu, Ou and Su, Fengguang and Deng, Yingjun},
  journal={ACM Transactions on Knowledge Discovery from Data},
  volume={18},
  number={4},
  pages={1--37},
  year={2024},
  publisher={ACM New York, NY}
}

@inproceedings{seedat2024dissecting,
  title={Dissecting Sample Hardness: A Fine-Grained Analysis of Hardness Characterization Methods for Data-Centric {AI}},
  author={Nabeel Seedat and Fergus Imrie and Mihaela van der Schaar},
  booktitle={The Twelfth International Conference on Learning Representations},
  year={2024},
  url={https://openreview.net/forum?id=icTZCUbtD6}
}

@article{lorena2018data,
  title={Data complexity meta-features for regression problems},
  author={Lorena, Ana C and Maciel, Aron I and de Miranda, P{\'e}ricles BC and Costa, Ivan G and Prud{\^e}ncio, Ricardo BC},
  journal={Machine Learning},
  volume={107},
  number={1},
  pages={209--246},
  year={2018},
  publisher={Springer}
}

@article{lorena2019complex,
  title={How complex is your classification problem? a survey on measuring classification complexity},
  author={Lorena, Ana C and Garcia, Lu{\'\i}s PF and Lehmann, Jens and Souto, Marcilio CP and Ho, Tin Kam},
  journal={ACM Computing Surveys (CSUR)},
  volume={52},
  number={5},
  pages={1--34},
  year={2019},
  publisher={ACM New York, NY, USA}
}

@article{leevy2018survey,
  title={A survey on addressing high-class imbalance in big data},
  author={Leevy, Joffrey L and Khoshgoftaar, Taghi M and Bauder, Richard A and Seliya, Naeem},
  journal={Journal of Big Data},
  volume={5},
  number={1},
  pages={1--30},
  year={2018},
  publisher={Springer}
}

@article{kaur2019systematic,
  title={A systematic review on imbalanced data challenges in machine learning: Applications and solutions},
  author={Kaur, Harsurinder and Pannu, Husanbir Singh and Malhi, Avleen Kaur},
  journal={ACM computing surveys (CSUR)},
  volume={52},
  number={4},
  pages={1--36},
  year={2019},
  publisher={ACM New York, NY, USA}
}

@inproceedings{tao2023dual,
  title={Dual focal loss for calibration},
  author={Tao, Linwei and Dong, Minjing and Xu, Chang},
  booktitle={International Conference on Machine Learning},
  pages={33833--33849},
  year={2023},
  organization={PMLR}
}

@inproceedings{he2008adasyn,
  title={ADASYN: Adaptive synthetic sampling approach for imbalanced learning},
  author={He, Haibo and Bai, Yang and Garcia, Edwardo A and Li, Shutao},
  booktitle={2008 IEEE international joint conference on neural networks (IEEE world congress on computational intelligence)},
  pages={1322--1328},
  year={2008},
  organization={IEEE}
}

@article{karras2022elucidating,
  title={Elucidating the design space of diffusion-based generative models},
  author={Karras, Tero and Aittala, Miika and Aila, Timo and Laine, Samuli},
  journal={Advances in neural information processing systems},
  volume={35},
  pages={26565--26577},
  year={2022}
}

@inproceedings{holte1989concept,
  title={Concept Learning and the Problem of Small Disjuncts.},
  author={Holte, Robert C and Acker, Liane and Porter, Bruce W and others},
  booktitle={IJCAI},
  volume={89},
  pages={813--818},
  year={1989}
}

@inproceedings{japkowicz2001concept,
  title={Concept-learning in the presence of between-class and within-class imbalances},
  author={Japkowicz, Nathalie},
  booktitle={Conference of the Canadian society for computational studies of intelligence},
  pages={67--77},
  year={2001},
  organization={Springer}
}

@article{jo2004class,
  title={Class imbalances versus small disjuncts},
  author={Jo, Taeho and Japkowicz, Nathalie},
  journal={ACM Sigkdd Explorations Newsletter},
  volume={6},
  number={1},
  pages={40--49},
  year={2004},
  publisher={ACM New York, NY, USA}
}

@inproceedings{guo2008class,
  title={On the class imbalance problem},
  author={Guo, Xinjian and Yin, Yilong and Dong, Cailing and Yang, Gongping and Zhou, Guangtong},
  booktitle={2008 Fourth international conference on natural computation},
  volume={4},
  pages={192--201},
  year={2008},
  organization={IEEE}
}

@article{ren2023systematic,
  title={A systematic review on imbalanced learning methods in intelligent fault diagnosis},
  author={Ren, Zhijun and Lin, Tantao and Feng, Ke and Zhu, Yongsheng and Liu, Zheng and Yan, Ke},
  journal={IEEE Transactions on Instrumentation and Measurement},
  volume={72},
  pages={1--35},
  year={2023},
  publisher={IEEE}
}

@inproceedings{mindermann2022prioritized,
  title={Prioritized training on points that are learnable, worth learning, and not yet learnt},
  author={Mindermann, S{\"o}ren and Brauner, Jan M and Razzak, Muhammed T and Sharma, Mrinank and Kirsch, Andreas and Xu, Winnie and H{\"o}ltgen, Benedikt and Gomez, Aidan N and Morisot, Adrien and Farquhar, Sebastian and others},
  booktitle={International Conference on Machine Learning},
  pages={15630--15649},
  year={2022},
  organization={PMLR}
}

@inproceedings{yuan2025instance,
  title={Instance-dependent Early Stopping},
  author={Suqin Yuan and Runqi Lin and Lei Feng and Bo Han and Tongliang Liu},
  booktitle={The Thirteenth International Conference on Learning Representations},
  year={2025},
  url={https://openreview.net/forum?id=P42DbV2nuV}
}

@article{ansuini2019intrinsic,
  title={Intrinsic dimension of data representations in deep neural networks},
  author={Ansuini, Alessio and Laio, Alessandro and Macke, Jakob H and Zoccolan, Davide},
  journal={Advances in Neural Information Processing Systems},
  volume={32},
  year={2019}
}

@inproceedings{pope2021intrinsic,
  title={The Intrinsic Dimension of Images and Its Impact on Learning},
  author={Phil Pope and Chen Zhu and Ahmed Abdelkader and Micah Goldblum and Tom Goldstein},
  booktitle={International Conference on Learning Representations},
  year={2021},
  url={https://openreview.net/forum?id=XJk19XzGq2J}
}

@article{ma2024unveiling,
  title={Unveiling and mitigating generalized biases of dnns through the intrinsic dimensions of perceptual manifolds},
  author={Ma, Yanbiao and Jiao, Licheng and Liu, Fang and Li, Lingling and Ma, Wenping and Yang, Shuyuan and Liu, Xu and Chen, Puhua},
  journal={IEEE Transactions on Pattern Analysis and Machine Intelligence},
  year={2024},
  publisher={IEEE}
}

@inproceedings{ma2023delving,
  title={Delving into Semantic Scale Imbalance},
  author={Yanbiao Ma and Licheng Jiao and Fang Liu and Yuxin Li and Shuyuan Yang and Xu Liu},
  booktitle={The Eleventh International Conference on Learning Representations },
  year={2023},
  url={https://openreview.net/forum?id=07tc5kKRIo}
}

@inproceedings{kienitz2022effect,
  title={The effect of manifold entanglement and intrinsic dimensionality on learning},
  author={Kienitz, Daniel and Komendantskaya, Ekaterina and Lones, Michael},
  booktitle={Proceedings of the AAAI Conference on Artificial Intelligence},
  volume={36},
  number={7},
  pages={7160--7167},
  year={2022}
}

@inproceedings{kaufman2023data,
  title={Data representations’ study of latent image manifolds},
  author={Kaufman, Ilya and Azencot, Omri},
  booktitle={International Conference on Machine Learning},
  pages={15928--15945},
  year={2023},
  organization={PMLR}
}

@inproceedings{ma2023curvature,
  title={Curvature-balanced feature manifold learning for long-tailed classification},
  author={Ma, Yanbiao and Jiao, Licheng and Liu, Fang and Yang, Shuyuan and Liu, Xu and Li, Lingling},
  booktitle={Proceedings of the IEEE/CVF conference on computer vision and pattern recognition},
  pages={15824--15835},
  year={2023}
}

@inproceedings{ahn2023cuda,
  title={{CUDA}: Curriculum of Data Augmentation for Long-tailed Recognition},
  author={Sumyeong Ahn and Jongwoo Ko and Se-Young Yun},
  booktitle={The Eleventh International Conference on Learning Representations },
  year={2023},
  url={https://openreview.net/forum?id=RgUPdudkWlN}
}

@article{vu2024lcsl,
  title={LCSL: long-tailed classification via self-labeling},
  author={Vu, Duc-Quang and Phung, Trang TT and Wang, Jia-Ching and Mai, Son T},
  journal={IEEE Transactions on Circuits and Systems for Video Technology},
  volume={34},
  number={11},
  pages={12048--12058},
  year={2024},
  publisher={IEEE}
}

@inproceedings{han2005borderline,
  title={Borderline-SMOTE: a new over-sampling method in imbalanced data sets learning},
  author={Han, Hui and Wang, Wen-Yuan and Mao, Bing-Huan},
  booktitle={International conference on intelligent computing},
  pages={878--887},
  year={2005},
  organization={Springer}
}

@article{dablain2022deepsmote,
  title={DeepSMOTE: Fusing deep learning and SMOTE for imbalanced data},
  author={Dablain, Damien and Krawczyk, Bartosz and Chawla, Nitesh V},
  journal={IEEE transactions on neural networks and learning systems},
  volume={34},
  number={9},
  pages={6390--6404},
  year={2022},
  publisher={IEEE}
}

@article{goodfellow2020generative,
  title={Generative adversarial networks},
  author={Goodfellow, Ian and Pouget-Abadie, Jean and Mirza, Mehdi and Xu, Bing and Warde-Farley, David and Ozair, Sherjil and Courville, Aaron and Bengio, Yoshua},
  journal={Communications of the ACM},
  volume={63},
  number={11},
  pages={139--144},
  year={2020},
  publisher={ACM New York, NY, USA}
}

@article{ho2020denoising,
  title={Denoising diffusion probabilistic models},
  author={Ho, Jonathan and Jain, Ajay and Abbeel, Pieter},
  journal={Advances in neural information processing systems},
  volume={33},
  pages={6840--6851},
  year={2020}
}

@article{ali2019mfc,
  title={MFC-GAN: Class-imbalanced dataset classification using multiple fake class generative adversarial network},
  author={Ali-Gombe, Adamu and Elyan, Eyad},
  journal={Neurocomputing},
  volume={361},
  pages={212--221},
  year={2019},
  publisher={Elsevier}
}

@article{zheng2020conditional,
  title={Conditional Wasserstein generative adversarial network-gradient penalty-based approach to alleviating imbalanced data classification},
  author={Zheng, Ming and Li, Tong and Zhu, Rui and Tang, Yahui and Tang, Mingjing and Lin, Leilei and Ma, Zifei},
  journal={Information Sciences},
  volume={512},
  pages={1009--1023},
  year={2020},
  publisher={Elsevier}
}

@article{marchesi2023generative,
  title={Generative AI Mitigates Representation Bias and Improves Model Fairness Through Synthetic Health Data},
  author={Marchesi, Raffaele and Micheletti, Nicolo and Kuo, Nicholas I-Hsien and Barbieri, Sebastiano and Jurman, Giuseppe and Osmani, Venet},
  journal={medRxiv},
  pages={2023--09},
  year={2023},
  publisher={Cold Spring Harbor Laboratory Press}
}

@article{doersch2016tutorial,
  title={Tutorial on variational autoencoders},
  author={Doersch, Carl},
  journal={arXiv preprint arXiv:1606.05908},
  year={2016}
}

@article{salimans2016improved,
  title={Improved techniques for training gans},
  author={Salimans, Tim and Goodfellow, Ian and Zaremba, Wojciech and Cheung, Vicki and Radford, Alec and Chen, Xi},
  journal={Advances in neural information processing systems},
  volume={29},
  year={2016}
}

@article{heusel2017gans,
  title={Gans trained by a two time-scale update rule converge to a local nash equilibrium},
  author={Heusel, Martin and Ramsauer, Hubert and Unterthiner, Thomas and Nessler, Bernhard and Hochreiter, Sepp},
  journal={Advances in neural information processing systems},
  volume={30},
  year={2017}
}

@inproceedings{szegedy2016rethinking,
  title={Rethinking the inception architecture for computer vision},
  author={Szegedy, Christian and Vanhoucke, Vincent and Ioffe, Sergey and Shlens, Jon and Wojna, Zbigniew},
  booktitle={Proceedings of the IEEE conference on computer vision and pattern recognition},
  pages={2818--2826},
  year={2016}
}

@inproceedings{wang2026difficulty,
  title={Difficulty Controlled Diffusion Model for Synthesizing Effective Training Data},
  author={Wang, Zerun and Mao, Jiafeng and Wang, Xueting and Yamasaki, Toshihiko},
  booktitle={Proceedings of the AAAI Conference on Artificial Intelligence},
  volume={40},
  number={12},
  pages={10367--10375},
  year={2026}
}

@inproceedings{vandenhende2019three,
  title={A three-player GAN: generating hard samples to improve classification networks},
  author={Vandenhende, Simon and De Brabandere, Bert and Neven, Davy and Van Gool, Luc},
  booktitle={2019 16th International Conference on Machine Vision Applications (MVA)},
  pages={1--6},
  year={2019},
  organization={IEEE}
}

@inproceedings{pennisi2021self,
  title={Self-improving classification performance through GAN distillation},
  author={Pennisi, Matteo and Palazzo, Simone and Spampinato, Concetto},
  booktitle={Proceedings of the IEEE/CVF International Conference on Computer Vision},
  pages={1640--1648},
  year={2021}
}

@article{ferracci2024targeted,
  title={Targeted synthetic data generation for tabular data via hardness characterization},
  author={Ferracci, Tommaso and Goldmann, Leonie Tabea and Hinel, Anton and Passino, Francesco Sanna},
  journal={arXiv preprint arXiv:2410.00759},
  year={2024}
}

@inproceedings{wang2023better,
  title={Better diffusion models further improve adversarial training},
  author={Wang, Zekai and Pang, Tianyu and Du, Chao and Lin, Min and Liu, Weiwei and Yan, Shuicheng},
  booktitle={International conference on machine learning},
  pages={36246--36263},
  year={2023},
  organization={PMLR}
}

@article{paszke2019pytorch,
  title={Pytorch: An imperative style, high-performance deep learning library},
  author={Paszke, Adam and Gross, Sam and Massa, Francisco and Lerer, Adam and Bradbury, James and Chanan, Gregory and Killeen, Trevor and Lin, Zeming and Gimelshein, Natalia and Antiga, Luca and others},
  journal={Advances in neural information processing systems},
  volume={32},
  year={2019}
}

\end{document}